\documentclass[sigconf, authorversion]{acmart}

\acmConference[ISSTA 2023]{ACM SIGSOFT International Symposium on Software Testing and Analysis}{17-21 July, 2023}{Seattle, USA}
\AtBeginDocument{%
  \providecommand\BibTeX{{%
    \normalfont B\kern-0.5em{\scshape i\kern-0.25em b}\kern-0.8em\TeX}}}

\usepackage{booktabs}
\usepackage{amsmath,amsfonts}
\newtheorem{definition}{Definition}
\usepackage{algorithmic}
\usepackage{graphicx}
\usepackage{subfigure}
\usepackage{textcomp}
\usepackage{framed}
\usepackage{bm}
\usepackage{enumerate}
\usepackage{tcolorbox}
\usepackage{xcolor}
\usepackage{arydshln}
\usepackage{multirow}
\usepackage{hyperref}
\usepackage[export]{adjustbox}
\usepackage{enumitem}

\setitemize[1]{itemsep=0pt,partopsep=0pt,parsep=\parskip,topsep=5pt}

\usepackage[ruled,linesnumbered]{algorithm2e}
\def\BibTeX{{\rm B\kern-.05em{\sc i\kern-.025em b}\kern-.08em
    T\kern-.1667em\lower.7ex\hbox{E}\kern-.125emX}}

\newcommand{\hz}[1]{{#1}}

\newcommand{\tool}[1]{\textsc{#1}\xspace}
\newcommand{\sysname}{\tool{FairRec}}
\newcommand{\drs}{DRSs}
\newcommand{\curps}{CurPS}
\newcommand{\infobase}{\textsc{InfoBase}\xspace}
\newcommand{\tabsysname}{\tool{FairRec}}
\newcommand{\rl}{\bm{rl}}

\begin{document}

\title{
\sysname: Fairness Testing for Deep Recommender Systems
}

\author{Huizhong Guo}
\affiliation{%
  \institution{Zhejiang University,Alibaba Group}
  \city{Hangzhou}
  \country{China}
}
\email{huiz_g@zju.edu.cn}

\author{Jinfeng Li}
\affiliation{%
  \institution{Alibaba Group}
  \city{Hangzhou}
  \country{China}
}
\email{jinfengli.ljf@alibaba-inc.com}

\author{Jingyi Wang}
\affiliation{%
  \institution{Zhejiang University}
  \city{Hangzhou}
  \country{China}
}
\email{wangjyee@zju.edu.cn}

\author{Xiangyu Liu}
\affiliation{%
  \institution{Alibaba Group}
  \city{Hangzhou}
  \country{China}
}
\email{eason.lxy@alibaba-inc.com}

\author{Dongxia Wang}
\authornote{Corresponding author.}
\affiliation{%
  \institution{Zhejiang University}
  \city{Hangzhou}
  \country{China}
}
\email{dxwang@zju.edu.cn}

\author{Zehong Hu}
\affiliation{%
  \institution{Alibaba Group}
  \city{Hangzhou}
  \country{China}
}
\email{zehong.hzh@alibaba-inc.com}

\author{Rong Zhang}
\affiliation{%
  \institution{Alibaba Group}
  \city{Hangzhou}
  \country{China}
}
\email{stone.zhangr@alibaba-inc.com}

\author{Hui Xue}
\affiliation{%
  \institution{Alibaba Group}
  \city{Hangzhou}
  \country{China}
}
\email{hui.xueh@alibaba-inc.com}

\begin{abstract}
Deep learning-based recommender systems (DRSs) are increasingly and widely deployed in the industry, which brings significant convenience to people's daily life in different ways. 
However, recommender systems are also shown to suffer from multiple issues, e.g., the \textit{echo chamber} and the \textit{Matthew effect}\footnote{The echo chamber refers to the effect that the user interest is reinforced by being repeatedly recommended similar items, resulting in a person being limited to a narrow range of content. The Matthew effect describes a phenomenon in which popular items have a cumulative advantage and therefore take up more exposure, limiting users' opportunities to see other items.}, of which the notation of ``fairness'' plays a core role. 
For instance, the system may be regarded as unfair to 
1) a certain user, if the user gets worse recommendations than other users,
or 2) an item (to recommend), if the item is much less likely to be exposed to the user than other items. 
While many fairness notations and corresponding fairness testing approaches have been developed for traditional deep classification models, they are essentially hardly applicable to DRSs. One major difficulty is that there still lacks a systematic understanding and mapping between the existing fairness notations and the diverse testing requirements for deep recommender systems, not to mention further testing or debugging activities.                
To address the gap, we propose \sysname, a unified framework that supports fairness testing of DRSs from multiple customized perspectives, e.g., model utility, item diversity, item popularity, etc.   
We also propose a novel, efficient search-based testing approach to tackle the new challenge, i.e., double-ended discrete particle swarm optimization (DPSO) algorithm, to effectively search for hidden fairness issues in the form of certain disadvantaged groups from a vast number of candidate groups. 
Given the testing report, by adopting a simple re-ranking mitigation strategy on these identified disadvantaged groups, we show that the fairness of DRSs can be significantly improved.    
We conducted extensive experiments on multiple industry-level DRSs adopted by leading companies. 
The results confirm that \sysname is effective and efficient in identifying the deeply hidden fairness issues, e.g., achieving $\sim$95\% testing accuracy with $\sim$half to 1/8 time.
\end{abstract}

\maketitle

\section{Introduction}
\label{sec:introduction}

Recommender systems (RSs) effectively bridge the gap between quality content and people who might be interested, benefiting and facilitating potential consumers and content providers in many fields, e.g., e-commerce, social media, etc.

\hz{A recommender system (RS) collects the feedback data of users in the real-time recommendation and updates itself on a periodic (e.g., daily) basis. 
In the dynamic process of data collection, model training or updating, item display, user clicks, etc., various biases may be introduced, which will subtly accumulate in the feedback-update loop of RSs. \cite{zhu2021popularity}.
\emph{If these biases are not disclosed in time, they could eventually lead to serious fairness issues, some of which are known as the \textit{echo chamber} and the \textit{Matthew effect.}}
For example, it is shown that in the MOOC platforms, courses taught by teachers from the United States are over-exposed by RSs, reducing the chances of teachers from the rest of the world being exposured~\cite{gomez2021winner}.
}

\hz{In recent years, with the rise of deep learning, deep learning-based recommender systems are increasingly and widely deployed in the industry, influencing millions of users in the world on a daily basis. 
For example, Microsoft released Deep Crossing \cite{shan2016deep} for advertisement recommendations for its products. Google released the Wide\&Deep \cite{cheng2016wide}, which was applied to application recommendations in Google play.
More recently, different types of deep recommendation models have been developed and introduced in both academia and industry \cite{guo2017deepfm, wang2017deep}.
In traditional statistical/collaborative filtering-based RSs, the correlation between the recommendation output and input features can be directly and explicitly obtained, which makes fairness issues easily detected by expert check.However, In DRSs, complicated DNN-based feature extraction and representation are unexplainable to humans. Therefore, it is challenging to analyze the deeply hidden fairness issues, and calls for a fine-grained fairness testing system for DRSs.
}

Fairness testing of deep learning models is an emerging research area in software engineering that aims to expose multiple kinds of fairness issues of a deep learning model, e.g., individual discrimination~\cite{kusner2017counterfactual}, group disparity~\cite{hardt2016equality}, etc., using various kinds of search strategies, e.g., random sampling \cite{galhotra2017fairness}, probabilistic sampling~\cite{udeshi2018automated}, gradient-based approach~\cite{zhang2020white} and symbolic execution~\cite{aggarwal2019black}. 
Despite the significant progress, existing fairness testing approaches are mostly for traditional deep classification problems and are essentially not applicable to DRSs, given the following critical challenges. 
\emph{First, there still lacks a systematic understanding and mapping between the fairness metrics from existing deep learning research~\cite{du2020fairness,wang2022providing} and the customized testing requirements of \drs~from a system point of view.
}
For instance, a DRS user may not only require accurate recommendation recommendation results but also diverse recommendation results. 
\emph{Second, the industry urgently calls for an efficient fairness testing approach which can be used to periodically evaluate their recommender systems and identify the discriminated user groups in time.}
The reason is that DRSs use multiple sensitive attributes of sparsely distributed users, forming a vast search space with a much more extensive collection of candidate groups (e.g., millions) than traditional fairness testing fields. 
Even worse, the large amount of user and behavior data makes it even more challenging to develop an efficient fairness testing algorithm for DRSs.

\begin{figure*}[t]
\setlength{\abovecaptionskip}{2pt}
\centerline{
\includegraphics[width=0.95\textwidth]{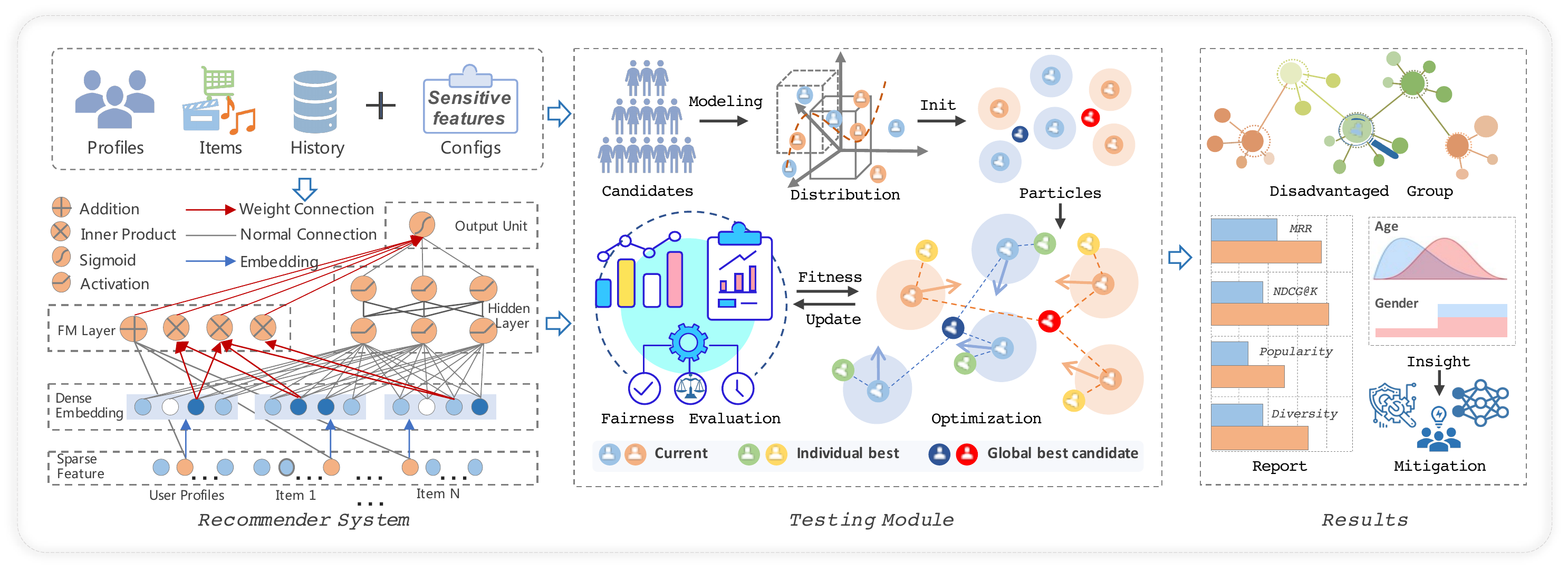}}
\caption{Our proposed \sysname framework.}
\label{fig:framework}
\vspace{-0.25cm}
\end{figure*}

In this work, we propose \sysname, a novel unified fairness testing framework specifically designed for DRSs to address the above challenges. 
As shown in Figure~\ref{fig:framework}, \sysname supports the testing of multi-dimensional fairness metrics such as model performance, diversity, and popularity, etc., meaning it considers not only how the model performs for different users but also how badly they might be influenced by user-tailored fairness issues such as the \textit{echo chamber} and the \textit{Matthew effect}, so as to meet the practical needs of different users on fair recommendation. 
Moreover, \sysname embraces a novel fairness testing algorithm based on the carefully designed double-ended discrete particle swarm optimization (DPSO) algorithm that can improve the testing efficiency by magnitude (compared to exhaustive search) while ensuring high testing accuracy.
Lastly, given the testing report of \sysname, by adopting some simple re-ranking mitigation strategy on those identified disadvantaged groups, the overall fairness of the DRSs can be significantly improved. 

In summary, we make the following contributions:

\begin{itemize}
\setlength{\leftmargin}{0.5em} 
\item We establish a systematic understanding and mapping between the diverse testing requirements and the existing fairness evaluation metrics to meet the various practical testing needs of DRSs.
\item We propose a novel fairness testing approach based on double-ended discrete particle swarm optimization (DPSO) algorithm which significantly improves the testing efficiency by magnitude while ensuring high testing accuracy. 
\item We evaluate \sysname with multiple industry-level deep recommendation models on popular large-scale datasets in this field. Our experiments confirm that \sysname is significantly more effective and efficient than the previous available methods, e.g., achieving $\sim$95\% testing accuracy with $\sim$half to 1/8 time.
\item We uncover the relation between the fairness performance w.r.t the different evaluation metrics in the experiments. 
\item We implement and release \sysname as an open source toolkit\footnote{\url{https://github.com/Grey-z/FairRec}
} together with all the data and models which could benchmark and facilitate future studies on testing and debugging of DRSs. 
\end{itemize}

\section{Preliminaries}
\label{sec:pre}

In this section, we briefly introduce the relevant background of DRSs and multi-attribute fairness, then formalize our problem.

\subsection{Deep Recommender System}\label{subsec:deep_recommender_system} 
A recommender system takes the historical behavior of users as input and outputs the items\footnote{We use ``item'' as an abstract concept to represent any content for the recommendation, e.g., news, jobs, goods, etc.} each user might be interested in. 
Let $\mathcal{U}$ be a user set and $\mathcal{V}$ be an item set, where $|\mathcal{U}|=n$, and $|\mathcal{V}|=m$. 
In this paper, we focus on deep learning-based recommender systems, which are usually more complicated and widely used in practice.
A DRS usually contain a recall layer and a ranking layer. 
For any user $u_i \in \mathcal{U}$, the recall layer will initially select $m'$ candidates from $\mathcal{V}$ based on the user's historical behavior. 
The ranking layer usually consists of a \textit{Deep Click-Through-Rate (CTR) Prediction model} which will estimate the user's preferences for all candidate items, and finally selects $k$ items to form the user's recommendation list $\rl_i$.

\begin{definition} \textbf{\textit{Deep recommender model}} \label{def:deep_recommender_system}
 Let $\bm{u}_i$ be the feature vector of user $u_i$, $\bm{u}_i = \left(\bm{s}_i,\bm{\hat{s}}_i\right)$ , where $\bm{s}_i$ denotes its sensitive attributes such as gender, age, and $\bm{\hat{s}}_i$ denotes its non-sensitive attributes.
 Let $\bm{v}_j$ be the feature vector of item $v_j$, the deep learning model is trained to predict the probability $p_{ij}$ that the user $u_i$ clicks the item $v_j$.
 A deep recommender model will then output $\rl_i$, which is a vector of $k$ items that have the top$-k$ highest probability values, as the recommendation result for each user $u_i$.
\end{definition}

\subsection{Fairness of Recommender System}\label{subsec:fairness_of_recsys}

There is a lack of consensus on the definition of fairness in tasks using deep learning.
From a high level, the most widely recognized concepts of fairness are individual fairness \cite{kusner2017counterfactual,grgic2016case} and group fairness \cite{hardt2016equality}. 
In personalized recommendation scenarios, individual fairness, which requires that two similar individuals should be treated similarly enough, is rather tricky to use.

In this work, we focus on group fairness which requires groups differing only in sensitive attributes to be treated with little distinction. 
Especially we focus on the fairness of groups formed by multiple sensitive attributes \cite{galhotra2017fairness}. 

\begin{definition}\textbf{\textit{Multi-attribute group}} \label{def:multi-attributes_group}
Let $l \in \mathbb{N}$ be the size of vector variable of the sensitive features $\bm{s}$. 
Let $X_{1}, X_{2} \ldots \ldots X_{l}$ be the possible sets of values of its entries $s_1 , s_2 \cdots ,s_l$ respectively. 
The set of multi-attribute groups defined by sensitive attributes can then be expressed as the Cartesian product $\mathcal{S}=X_1 \times X_2 \times \cdots \times X_{l}$, an element of which describes a specific user group e.g., group $\bm{G_{s_j}}$ with $\bm{s_j}=(x_{j1}, x_{j2},\cdots,x_{jl}), \bm{s_j} \in \mathcal{S}$. 
\end{definition}

Let $\mathcal{M}\left(u_i,\rl_i \right)$ represents the performance value regarding an evaluation metric of a recommendation result for a user.
Given two multi-attribute groups and a metric in concern, we use the following absolute value to measure the distance between the performance of the recommendations they receive, i.e.,
\begin{equation}
\setlength{\abovedisplayskip}{-1ex}
\nonumber
\footnotesize
MGD_{\mathcal{M}}(\bm{G_{s_1}},\bm{G_{s_2}}) \!=\! \lvert\frac{1}{\left|\bm{G_{s_1}}\right|} \sum_{u_i \in \bm{G_{s_1}}}\!\mathcal{M}\left(u_i,\rl_i
\right)\!-\!\frac{1}{\left|\bm{G_{s_2}}\right|} \sum_{u_j \in \bm{G_{s_2}}}\!\mathcal{M}\left(u_j,
\rl_j
\right)\lvert ,
\end{equation}
which enables us to define multi-attribute group fairness.

\begin{definition}
\textbf{\textit{Multi-attribute group fairness}} \label{def:multi-attributes_group_fairness}
For the entire user population, we use the maximum gap w.r.t. an evaluation metric $\mathcal{M}$ between two groups to measure the group fairness of a DRS, which has been widely adopted in previous work \cite{perera2022search,galhotra2017fairness}. Given any metrics $\mathcal{M}$, the unfairness of a DRS can be defined as follows:
\end{definition}
\begin{equation}
\label{eq:fairness_metric}
    UF_\mathcal{M} = \max _{\forall i, j} MGD_{\mathcal{M}}(\bm{G_{s_i}},\bm{G_{s_j}}).
\end{equation}
$UF_\mathcal{M}$ measures the degree of group unfairness of a DRS w.r.t an evaluation metric $\mathcal{M}$.
The larger the value, the more unfair a DRS is w.r.t the metric.
The value itself and also the user groups that correspond to the value form important parts of our testing results.

\subsection{Problem Formalization}
\label{sec:pf}
Fairness testing refers to any activity designed to reveal fairness bugs \cite{chen2022fairness}. 
For a given DRS $\mathcal{R}$ that recommends $k$ items from the item set for each user $i$, the primary goal of \sysname is to \emph{1) measure the multi-attribute group fairness score} defined above. 
Note that the reported score $UF_\mathcal{M}$ is parameterized by a performance evaluation metric $\mathcal{M}$, which allows \sysname to measure the system fairness from different perspectives.
Along with the measurable fairness score, \emph{2) \sysname also reports a certain user-defined number of advantaged user groups and disadvantaged groups}, which could facilitate the model developer to further debug the model and improve the model's group fairness. 
To achieve this goal, we need to find the testing candidates (user groups) for which the performance values regarding a metric \emph{differ as much as possible}, such that we can accurately uncover the severity of unfairness problem of a DRS.

\section{The \sysname Framework}
\label{sec:fr}
In this section, we first provide an overview of our testing framework, and then present the detailed metrics and testing approach.

\subsection{System Overview}
\label{subsec:system_overview}
We present the framework of \sysname in Figure~\ref{fig:framework}, which consists of three main components, i.e., the input module, the testing module and the results display module.

Concretely, in the \textbf{input module}, the user and item data, as well as the recommendation models, should be included.
Notice that \sysname is a model-agnostic testing system, it only requires black-box queries access to the target recommender models.
In addition, the testing requirement of interest such as the sensitive user attributes $S$ (e.g., gender and age) and fairness metrics $\mathcal{M}$ of interest should also be configured. In the \textbf{testing module}, \sysname first loads the data and models and then identifies the advantaged and disadvantaged groups concerning the specified fairness metric via our specifically designed double-ended discrete particle swarm optimization algorithm according to the configured requirements.
Specifically, more details of our designed effective and efficient search-based testing algorithm will be included in Section~\ref{subsec:search_based_testing_with_dpso}. Finally in the \textbf{display module}, \sysname displays a multi-dimensional testing report that includes the overall fairness evaluation results as well as the details of the found disadvantaged groups, aiming to provide insight on the bias mitigation followed.

\subsection{RS-Tailored Evaluation Metrics} \label{subsec:rs_evaluation_metrics}
Group fairness needs to be measured in terms of an evaluation metric (EM) for traditional deep classification tasks or deep recommendation tasks. 
However, different from the classification tasks where the commonly agreed evaluation metric is the prediction accuracy, it is more than challenging to define `accuracy' for a personalized recommendation task given the flexibility, diversity, and subjectivity of user's practical needs. 
In fact, how to define the metrics which can properly reflect these aspects is still popular ongoing research in the RS community~\cite{yao2017beyond,fairmetric2022wang}. 

In this work, we focus on fairness from a user's perspective and systematically adopt five RS-tailored evaluation metrics in \sysname. Note that \sysname is flexible to incorporate more evaluation metrics later. 
These metrics are selected considering multiple practical needs, i.e., 1) model performance metrics (AUC, MRR and NDCG), 2) diversity metric (URD to measure the \textit{echo chamber} effect), and 3) popularity metric (URP to measure the impact of \textit{Matthew effect} on users).
\hz{In the following, we elaborate the details of these metrics in the top-k recommendation.}

\textbf{EM1.1 \emph{Area under Curve (AUC)}}~\cite{ling2003auc}. It is defined as the area under the receiver operating characteristic curve, which is plotted with the true positive rate as the vertical coordinate and the false positive rate as the horizontal coordinate. 
AUC is the most commonly used metric to measure the performance of DRSs.
AUC measures how likely an item of interest is ranked higher than another out of interest in a recommended list.  
A higher AUC value means a DRS recommends more accurately.

\textbf{EM1.2 \emph{Mean Reciprocal Rank (MRR)}} \cite{voorhees2001trec}. 
It measures on average, how top user-target items are ranked in their recommended lists.
MRR is calculated with reciprocal ranks as below:

\begin{equation}
\label{eq:mrr}
    MRR = \frac{1}{|\bm{G_{s}}|}\sum_{u_i\in \bm{G_{s}},v_j \in \mathcal{V}}\frac{1}{pos(v_j|r_{ij}=1)},
\end{equation}
where $r_{ij} \!=\! 1$ denotes that the item $v_j$ is the target item of $u_i$ and $pos(v_j)$ denotes the ranking position of $v_j$ in the recommendation list of $u_i$. If for a user, there is no target item in his list, then the fraction in the sum is assigned $0$. 
A higher MRR value means the target items are ranked higher (which can lead to more clicks), indicating more accurate recommendation. 

\textbf{EM1.3 \emph{Normalized Cumulative Gain for k Shown Recommendations (NDCG@k)}}~\cite{jarvelin2002cumulated}. 
It is a commonly used metric for evaluating the performance of a recommender system based on the graded relevance of the recommended items, which varies from 0.0 to 1.0, with 1.0 representing the ideal recommendation result, i.e., users get the recommendations they are interested in and their favorite items get the most exposure.  
\begin{equation}
\setlength{\abovedisplayskip}{4pt}
\label{eq:dcg}
    D C G @ k=\sum_{a=1}^{K} \frac{2^{p(v_a)}-1}{\log _{2}(a+1)},
\end{equation}

\begin{equation}
\label{eq:ndcg}
    N D C G_{u} @ k=\frac{D C G_{u}}{I D C G_{u}},
\end{equation}
where $a$ denotes the sorting position of the item, and $p(v_a)$ denotes the model prediction output of $v_a$. 
$DCG_u$ and $IDCG_u$ are calculated based on the recommended list and the ideal list of $u_i$, respectively. 
A higher value of (NDCG@k) indicates better recommendation.

It can be noticed that in general, the metrics AUC, MRR and NDCG@k all describe how well the recommendations match the interests of users.
However, sometimes a user may not only want to be shown the items that he already knows he would be interested in, but also want to see other types of items.
We need recommendation diversity metric to capture this.

\textbf{EM2 \emph{Diversity (URD)}}~\cite{qin2013promoting}. 
\hz{It measures the level of diversity in recommendations for users. 
A higher value indicates that users receive a more varied set of recommendations, thereby reducing the impact of the \textit{echo chamber} effect.}
Specifically, we use the \textit{intra-list similarity} \cite{ziegler2005improving} to calculate the diversity of a recommendation list, 

\begin{equation}
\setlength{\abovedisplayskip}{-1ex}
\label{eq:ils}
    URD(\rl_i)=1 - \frac{2}{k(k-1)} \sum_{v_a \in \rl_i } \sum_{v_b \neq v_a \in \rl_i} \operatorname{Sim}(v_a, v_b),
\end{equation}
where $a$, $b$ denotes the sorting position of the item in $\rl_i$, $v_a$ denotes the $a$-th item and $Sim(v_a,v_b)$ denotes the similarity between $v_a$ and $v_b$. 
In this work, we use the Jaccard similarity to calculate the diversity of the recommended items. 

\textbf{EM3 \emph{Popularity (URP)}}~\cite{abdollahpouri2021user}. 
\hz{It measures how well the popularity of the recommended items matches that of the user’s preference.
A higher value indicates the user is more affected by the Matthew effect, e.g., a user prefers niche movies but only be recommended with very popular ones.}
How popular an item $v_j$ is in general is calculated by
\begin{equation}
\label{eq:rp}
   RP_{j} = \frac{\phi(j)}{|\mathcal{T}|} * 100\% ,
\end{equation}
where $\mathcal{T}$ denotes the training set and $\phi(j)$ denotes the number of times item $j$ has been interacted in the training set. 
Based on Equation.~\ref{eq:rp}, we use the following absolute value to define the URP metric.
The smaller the value is, the better the popularity of the recommended items matches the popularity preference of the user. 

\begin{equation}
\setlength{\abovedisplayskip}{-1ex}
\label{eq:urp}
    URP(\rl_i,\bm{hl}_i) =  \frac{1}{|\rl_i|} \sum_{v_j \in \rl_i} RP_{j}-\frac{1}{|\bm{hl}_i|} \sum_{v_j \in \bm{hl}_i} RP_{j} ,
\end{equation}
where $\bm{hl}_i$ and $\bm{rl}_i$ denote the historical interaction list and the recommendation list of $u_i$, respectively. 

\subsection{Search-based Testing Algorithm}
\label{subsec:search_based_testing_with_dpso}

Given a specific evaluation metric, the core challenge of multi-attribute group fairness testing is to measure the maximum gaps between the most advantaged and disadvantaged user groups.
This is highly non-trivial for realistic DRSs due to the following challenges:
(i) First, compared to traditional deep learning fairness testing literature which only target a single or few sensitive attributes, \emph{there might be dozens of sensitive attributes (e.g., gender, age, nationality, etc.) with multi-dimensional discrete values in a DRS}.
The combinations of these attributes maps the testing candidates into an extremely high-dimensional and sparse search space (e.g., the number of user groups to search for could be in millions);
(ii) Even worse, in an industrial DRS, \emph{there can be a huge amount of users (e.g., 360K in some of our experiment) who are sparsely distributed in the search space}. The complexity of testing such a real DRS inherently requires extremely high testing budget while it is often required to be completed within an acceptable limited time, making high testing efficiency critical.
 
In \sysname, we propose a novel double-ended discrete particle swarm optimization (DPSO) algorithm to address the challenging search problem for a DRS and make the testing much more efficient while ensuring high accuracy.
The Particle swarm optimization (PSO) algorithm is an evolutionary computation technique \cite{kennedy1995particle} to search for the optimal solution of an optimization problem, derived from the simulation of the social behavior of birds within a flock. Previously, PSO has been shown to be effective and efficient in solving multiple kinds of testing problems \cite{lyu2019mopt, windisch2007applying}.

Algorithm~\ref{algo:dpso} shows the overall testing workflow of \sysname. The inputs include the test dataset $D$, the set of sensitive attribute $S$, and the testing budget, i.e., the maximum number of iteration $n$. 
At line 1, DPSO first sets up the search space, the optimization objective of testing and initialize the particle swarms. From line 2 to line 6, DPSO will iteratively evaluate the testing objective and update the particle swarms until the testing budget is exhausted.
Several important customization and optimizations are particularly made for the fairness testing problem of DRS. 

\begin{algorithm}[t]
\caption{$DPSO$ of \sysname}
\label{algo:dpso}
\KwIn{Test Dataset $D$, Sensitive Attributes $\bm{s}$, Maximum iterations $n_{it}$}
\KwOut{Fairness score $UF_\mathcal{M}$, target groups $\bm{G_{s_a}}, \bm{G_{s_d}}$ }
$\curps_a , \curps_d = Initialize(D,\bm{s})$\; 
$i = 0$ \;
\While{$i < n_{it}$}{
$\curps_a , \curps_d, \infobase = Update(\curps_a , \curps_d)$\;
$i++$\;
}
$\bm{G_{s_a}} = max(\infobase)$\;
$\bm{G_{s_d}} = min(\infobase)$\;
$UF_\mathcal{M} = \bm{G_{s_a}} - \bm{G_{s_d}}$ \;
\Return{ $\bm{G_{s_a}}, \bm{G_{s_d}}$, $UF_\mathcal{M}$}
\end{algorithm}

First, to effectively evaluate the maximum gap between two user groups, we propose a bi-end search solution with two particle swarms running simultaneously, where one particle swarm is designed to find the most advantaged group $\bm{G_{s_{a}}}$ and the other swarm is designed to find the most disadvantaged group $\bm{G_{s_{d}}}$ in an opposite direction according to Equation~\ref{eq:fairness_metric}. 
Specifically, each particle represents a user group segmented by $l$ sensitive attributes, which is a point in the $l$-dimensional search space.
For instance, the group consisting of 30-year-old male doctors (i.e., \textit{gender}=male/0, \textit{age}=30 and \textit{occupation}=doctor/9) can be denoted as the point $(0, 30, 9)$ in the 3-dimension search space. 

Second, during the testing process, each particle flies through the search space and memorizes its individual best position $Pbest$ (i.e., the grouping condition that results in a current maximum fairness gap between the divided groups), which is shared with the other particles.
The global optimal grouping condition $Gbest$ can then be obtained from the $Pbest$ of the two swarms (with different goals in mind) from the whole search history. 
In each search iteration, particles in the two swarms share better positions with each other and dynamically adjust their own position and velocity in each dimension according to the calculated \textit{fitness} (i.e., the difference between the metrics of the two divided groups) of its own and other members.
The whole search process is iterated until it finally converges to the global optimal position $Gbest$, and the difference in the group-fairness-metric values $MGD_{\mathcal{M}}(\bm{G_{s_{a}}},\bm{G_{s_{d}}})$ (Refer to Equation~\ref{eq:fairness_metric}) is reported as the fairness measurement of the DRS. Note that the algorithm can be easily configured to memorize a certain number of disadvantaged groups for the final report through maintaining a list of $Pbest$ for them during the search process.

We then introduce the carefully designed optimizations of the two key steps in DPSO to further improve the testing efficiency.

\begin{figure}
    \setlength{\abovecaptionskip}{2pt}
    \setlength{\belowcaptionskip}{1pt}  
    \centering
    \subfigure[Uniform-based]{
        \centering
        \includegraphics[width=0.2\textwidth]{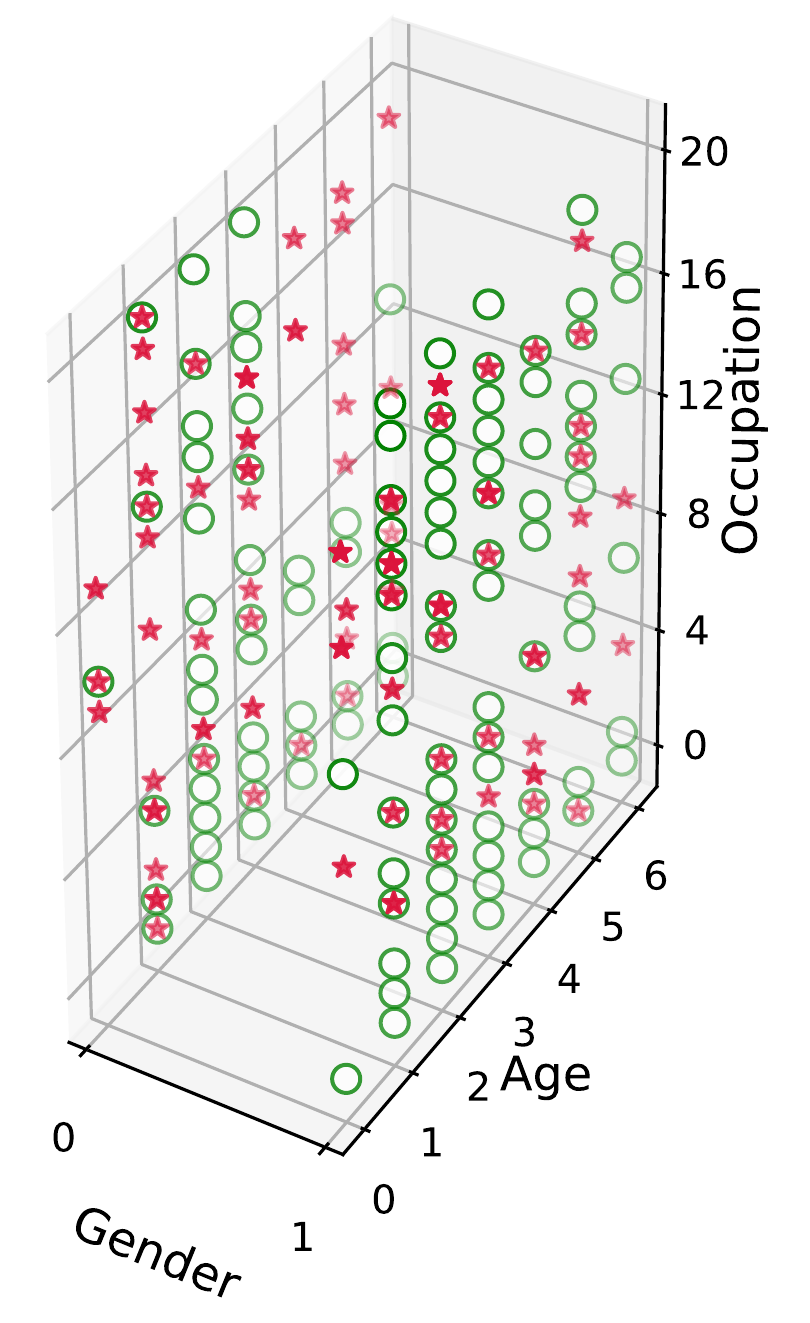}
        \label{fig:initial_random_uniform}
    }
    \subfigure[Distribution-based]{
        \centering
        \includegraphics[width=0.2\textwidth]{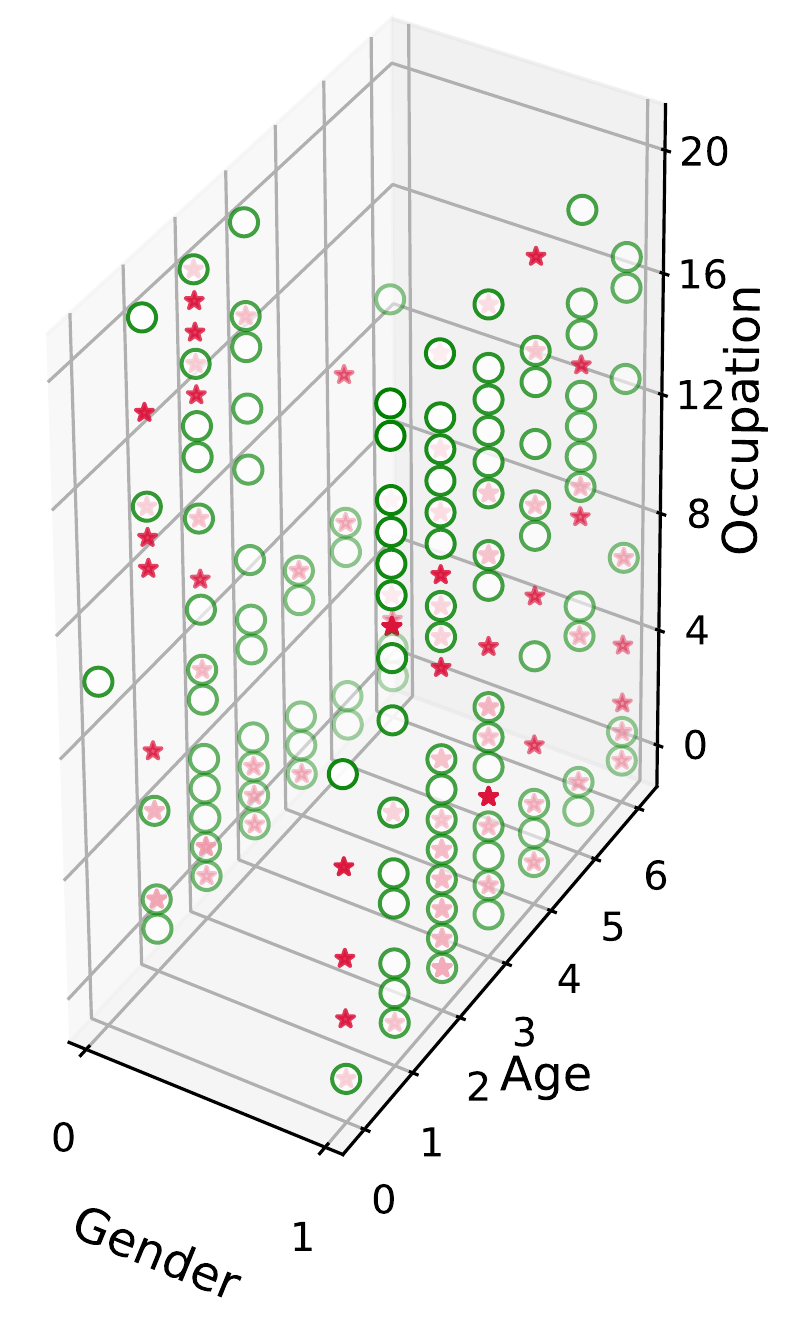}
      \label{fig:initial_distribution}
    }
    \caption{Comparison of different initialization methods.}
    \label{fig:initial_comparison}
    \vspace{-0.5cm}
\end{figure}

\textbf{Initialize.} 
As mentioned above, users are sparsely distributed in the multi-dimensional partitioned feature space. 
We therefore propose to initialize the two particle swarms according to the actual distribution of the testing candidate users to quickly locate the target area in the search space.
The concrete initialization scheme is illustrated in Algorithm~\ref{algo:initialize}.
Firstly, we set the search boundary $B^{s_i}$ (line 1) and the probability distribution $D^{s_i}$ (line 2) of each sensitive attribute $s_i \in \bm{s}$.
For example, given $s_i$ = ``gender", assume a group contains 4 females $(s_i = 0)$ and 6 males $(s_i = 1)$, we have $B^{s_i} = [0, 1]$ and $D^{s_i} = [0.4, 0.6]$.
Then, we initialize the particle swarms based on the distribution of users, where $\curps_a$ and $\curps_d$ denote the particle swarms with search goals of $\bm{G_{\bm{s}_a}}$ and $\bm{G_{\bm{s}_d}}$, respectively (line 3-4).
Finally the velocity $V$ and individual best $Pbest$ of every particle $P$ are initialized (line 5-8).
We also visualize the distribution-based initialization method we proposed and the traditional initialization method in Figure~\ref{fig:initial_comparison} for a more intuitive comparison, in which the green circles represent the actual users distribution in testing candidates. 
The red stars in Figure~\ref{fig:initial_random_uniform} denote the particles initialized via random initialization, while the red stars in Figure~\ref{fig:initial_distribution} are the particles initialized in \sysname based on the modeled distribution. 
It can be seen from Figure~\ref{fig:initial_comparison} that the distribution-based initialization we proposed significantly improves the particle coverage, which allows the generated particles to be more concentrated on the target region, thus improving the effectiveness of searching for the two target groups.

\begin{algorithm}[t]
\caption{ Initialize }
\label{algo:initialize}
\KwIn{Test Dataset $D$, Sensitive Attributes $\bm{s}$, Size $n_{pt}$}
\KwOut{$ IniPS_a, IniPS_d $}
$B^{s_i} = SetBoundary(D,\bm{s})$\;
$D^{s_i} = GetDistribution(D,\bm{s})$ \;
$IniPS_a = Sample(B^{s_i}, D^{s_i}, n_{pt})$ \;
$IniPS_d = Sample(B^{s_i}, D^{s_i}, n_{pt})$ \;
$IniPS = IniPS_a \cup IniPS_d $ \;
\ForEach{particle $P_i$ in $IniPS$}{
$Vi = random.uniform(-v^*,v^*)$\;
$Pbest_i \leftarrow P_i $\;
}
\Return{$ IniPS_a, IniPS_d $}
\end{algorithm}

\begin{algorithm}
\caption{Fairness evaluate and update}
\label{algo:fairness_evaluate_and_update}
\KwIn{$ \curps_a, \curps_d$}
\KwOut{$ \curps_a, \curps_d$}
$ \infobase \leftarrow \varnothing$ \;
\ForEach{particle $P_i$ in $\curps_a \cup \curps_d$}{
$P^{k+1}_i, V^{k+1}_i = Update(P^k_i,V^k_i,Pbest_i,Gbest_i)$\;
\If{$P_i$ in \infobase}{
$fitness(P_i) \leftarrow Retrieve(\infobase)$\;
}
\Else{
$fitness(P_i) = FairnessEvaluate(P_i))$\;
$ \infobase \leftarrow \infobase \cup fitness(P_i)$ \;
}
\If{$P_i \in \curps_a$}{
$Pbest_i \leftarrow max\{fitness(Pbest_i),fitness(P_i)\} $\;
}
\If{$P_i \in \curps_d$}{
$Pbest_i \leftarrow min\{fitness(Pbest_i),fitness(P_i)\} $\;
}
}
\Return{$ \curps_a, \curps_d, \infobase $}
\end{algorithm}

\textbf{Update.} 
Algorithm~\ref{algo:fairness_evaluate_and_update} illustrates the details of fairness evaluation and update. 
After initialization, all particles will move toward the target groups at each iteration, guided by $Pbest$ and $Gbest$ (line 3).
The velocity and position of all particles is updated according to the formula defined as follows,

\begin{equation}
\setlength{\abovedisplayskip}{-1ex}
\setlength{\belowdisplayskip}{2pt}
    V_i^{k+1} = \alpha \cdot |C^k - P_i^k| \cdot \varphi_i + c_1 \cdot r_1 \cdot(Pbest_i^k - P_i^k) + c_2 \cdot r_2 \cdot(Gbest^k - P_i^k)
\label{eq:update_velocity}
\end{equation}

\begin{equation}
\setlength{\abovedisplayskip}{2pt}
\setlength{\belowdisplayskip}{2pt}
     P_i^{k+1} = P_i^k + V_i^{k+1},
\label{eq:update_position}
\end{equation}
where $P_i^{k}$ and $V_i^{k}$ are the position and velocity of i-th particle at the k-th iteration, $r_1$ and $r_2$ are random numbers between 0 and 1, and $\varphi_i$ is a random number that obeys the standard normal distribution.
Each particle $P_i$ will update the next search direction under the dual guidance of the individual best position $Pbest_i$ and the global best position $Gbest$.
Constants $c_1$ and $c_2$ are the weighting factors of the stochastic acceleration terms, which pull particle $P_i$ towards $Pbest_i$ and $Gbest$ positions.
To improve the ability of the particles to escape from the local optimal position, we introduce the thermal motion~\cite{sun2011solving}, as the first term of the Equation.~\ref{eq:update_velocity}, where $\alpha$ denotes the inertia factor which controls the magnitude of the thermal motion.

After each particle updates its position, the fairness fitness will be calculated and stored in the \infobase (line 8-9).
The \infobase provides a platform for all particles to share information, thus avoiding duplicate computation (line 4-6).
After each iteration, $Pbest_i$ of each particle $P_i$ will be updated (line 11-16) and the two swarms update the candidate global best position $Gbest$ using the shared information (line 17-18).
Notice that the update directions of the two swarms are opposite since the search targets are different.

\section{Experiment}
\label{sec:exp}

In this section, we first describe our experimental setup, and then evaluate the effectiveness and efficiency of \sysname by answering several key research questions.

\subsection{Experimental Setup}

\begin{table}[t]
\setlength{\abovecaptionskip}{6pt}
\setlength\tabcolsep{3pt}
\centering
\caption{Recommendation performance of the models.}
\label{tab:details_of_trained_model}
\resizebox{0.95\linewidth}{!}{
\begin{tabular}{*{9}{ccccccccc}}
  \toprule
   \multirow{2}{*}{\textbf{Model}}    & \multicolumn{2}{c}{\textbf{MovieLens}} &  \multicolumn{2}{c}{\textbf{LFM360K}}  
       & \multicolumn{2}{c}{\textbf{BlackFriday}} &  \multicolumn{2}{c}{\textbf{Amazon}} 
  \\
  \cmidrule{2-9}
   & loss & auc & loss & auc  & loss & auc & loss & auc
  \\
  \midrule
  \multirow{1}{*}{Wide\&Deep} & 0.3865 &	0.9143 & 0.3737 & 0.9219 & 0.2732 & 0.9600 & 0.3788 & 0.9298
  \\
  \midrule
  \multirow{1}{*}{DeepFM} & 0.3791 & 0.9188 & 0.3767 & 0.9210 & 0.2726 &	0.9602 & 0.3709 & 0.9331
  \\
  \midrule
  \multirow{1}{*}{DCN} & 0.3797 & 0.9183 & 0.3739 & 0.9220 & 0.2744 & 0.9597 & 0.3881 & 0.9259
  \\
  \midrule
  \multirow{1}{*}{FGCNN} & 0.3842 & 0.9211 &	0.3719 & 0.9290 & 0.3020 & 0.9569 &	0.3278 & 0.9587
  \\
  \bottomrule
\end{tabular}}
\vspace{-0.35cm}
\end{table}

\textbf{Datasets.}
We adopt four widely used large-scale public recommendation benchmark datasets in our experiments, the details of these datasets are as follow:
{\textbf{\emph{MovieLens}}~\cite{harper2015movielens} is composed of 1 million movie rating data from 6,040 users for 3,883 movies.
The sensitive attributes in this dataset are gender, age, and occupation, which can divide users into 294 groups.}
{\emph{\textbf{LFM360K}}~\cite{celma2009music} contains music listening history data from 360,000 users with sensitive attributes of gender, age and nationality.
In our experiment, a interaction data of 260,000 users and 140,000 music tracks is reserved for training after filtering out those data with missing values, which can be divided into 53,058 groups based on sensitive attributes.}
{\emph{\textbf{BlackFriday}}~$\footnote{https://www.kaggle.com/datasets/sdolezel/black-friday}$ is a sales dataset obtained from the Kaggle website which consists of 550,068 purchases from 5,889 users for a total of 3,566 items. 
The sensitive attributes in this dataset are gender, age, occupation, city category, years of stay in the current city, and marital status, based on which users can be divided into 8,820 groups.}  
{\emph{\textbf{Amazon}}~\cite{wang2021deconfounded} electronics dataset includes purchase behavior data from 175,878 users with 944,347 reviews for 29,391 items on Amazon.
It does not contain any personal user information.
Referring to the setup of \cite{li2021user}, we categorize users based on their historical behavior regarding the number of interactions, total consumption, and the highest price of items purchased, and set them as the sensitive attributes. 
Each attribute contains ten values from 0 to 9, thus users can be divided into 1,000 groups.} 

Following the setting of~\cite{he2017neural}, we divide the last positive interaction data of users into the test set and the remaining into the training set. 
For each positive interaction, we randomly sample one item without user interaction as a negative interaction in the training set. 
We further randomly sample 49 items that users have not interacted with and combine them with the target item to form the candidate items for recommendation during testing.

\textbf{Recommendation Models.}
We adopt four classic deep CTR prediction models for the recommendation, i.e., Wide\&Deep~\cite{cheng2016wide}, DeepFM~\cite{guo2017deepfm}, DCN~\cite{wang2017deep} and FGCNN~\cite{liu2019feature}, owing to the exhaustive academic efforts on it and tremendous amount of industry applications.
These models are trained on the above four datasets following the setup of ~\cite{shen2017deepctr}.
In the testing stage, the top-5 recommendation list is generated from the 50 candidate items according to the predicted CTR in the descending order.
The evaluated recommendation performance of the reproduced models are shown in Table~\ref{tab:details_of_trained_model}.

\begin{table*}[tp]
\setlength{\abovecaptionskip}{6pt}
\setlength\tabcolsep{8pt}
\centering
\caption{The main results of fairness testing from the perspective of effectiveness and efficiency.}
\label{tab:effectiveness_result}
\resizebox{0.835\textwidth}{!}{
\begin{tabular}{*{14}{cccccccrcccccr}}
  \toprule
      &  & \multicolumn{6}{c}{\textbf{MovieLens}}  &  \multicolumn{6}{c}{\textbf{LFM360K}} 
  \\
  \cmidrule(lr){3-8} \cmidrule(lr){9-14}
    & & \multicolumn{1}{c}{$UF_{auc}$} & \multicolumn{1}{c}{$UF_{mrr}$} & \multicolumn{1}{c}{$UF_{ndcg}$} & \multicolumn{1}{c}{$UF_{urd}$} & \multicolumn{1}{c}{$UF_{urp}$} & \multicolumn{1}{c}{Time(s)} &
    \multicolumn{1}{c}{$UF_{auc}$}&
    \multicolumn{1}{c}{$UF_{mrr}$} & \multicolumn{1}{c}{$UF_{ndcg}$} & \multicolumn{1}{c}{$UF_{urd}$} & \multicolumn{1}{c}{$UF_{urp}$}  & \multicolumn{1}{c}{Time(s)}
  \\
  \midrule
  \multirow{4}{*}{\textbf{Wide\&Deep}} & Themis & 0.2537 & 0.4819 & 0.1719 & 0.1786 & 0.0601 & 140.53
  & 0.0655 & 0.2049 & 0.0639 & 0.00045 & 0.0114 & 16276.26
  \\
  \cdashline{3-14}[2pt/2pt]
  & TestSGD$_{\theta=0.01}$ & 0.0683 & 0.1978 & 0.0597  & 0.0467 & 0.0148  & 58.65
  
  & 0.0571 & 0.1406 & 0.0443  & 0.00014 &  0.0031 &  2671.80
  \\
  & TestSGD$_{\theta=0.005}$ & 0.0861 & 0.2229 & 0.0720  & 0.0655  & 0.0274 & 82.63
  
  & 0.0574 & 0.1715  &  0.0560   & 0.00024  &  0.0072  &  2805.50
  \\
    
   & \tabsysname & \textbf{0.2436} & \textbf{0.4603} & \textbf{0.1622} & \textbf{0.1605} & \textbf{0.0592} & 86.95
   
   & \textbf{0.0645} & \textbf{0.2043} & \textbf{0.0628} & \textbf{0.00041} & \textbf{0.0105} & 2121.41

  \\
  \midrule
  \multirow{4}{*}{\textbf{DeepFM}} & Themis & 0.2341 & 0.6226 & 0.2009 & 0.1767 & 0.0623 & 117.41
  & 0.0627 & 0.2119 & 0.0639 & 0.00054 & 0.0115 & 16270.25
  \\
  \cdashline{3-14}[2pt/2pt]
   & TestSGD$_{\theta=0.01}$ & 0.0667 & 0.2052 & 0.0636  & 0.0646 & 0.0157 & 62.72
   
  & 0.0598 & 0.1383 & 0.0433  & 0.00010 & 0.0045 & 2698.31
  \\
  & TestSGD$_{\theta=0.005}$ & 0.0849 & 0.2500 & 0.0813  & 0.0728 & 0.0220 & 99.36
  
  & 0.0601 & 0.1700 & 0.0547 & 0.00031 & 0.0072 & 2895.77
  \\ 
    
   & \tabsysname & \textbf{0.2341} & \textbf{0.5772} & \textbf{0.2009} & \textbf{0.1646} & \textbf{0.0594} & 99.06
   
   & \textbf{0.0627} & \textbf{ 0.2103} & \textbf{0.0638} & \textbf{0.00051} & \textbf{0.0115} & 2151.46
  \\
  \midrule
  \multirow{4}{*}{\textbf{DCN}} & Themis & 0.2514 & 0.5185 & 0.1768 & 0.1528 & 0.0637 & 112.28
  
  & 0.0632 & 0.2118 & 0.0656 & 0.00053 & 0.0113 & 16303.43

  \\
  \cdashline{3-14}[2pt/2pt]
  & TestSGD$_{\theta=0.01}$ & 0.0699 & 0.1493 & 0.0572 & 0.0516 & 0.0168  & 50.33
  
  &  0.0586 & 0.1410 & 0.0446 & 0.00011 & 0.0037 & 2788.76
  \\
  
  & TestSGD$_{\theta=0.005}$ & 0.0907 & 0.2071 & 0.0708  & 0.0793 & 0.0298 & 103.93
  
  & 0.0599 & 0.1687 & 0.0537 & 0.00026 & 0.0073 & 2930.42
  \\
    
  & \tabsysname & \textbf{0.2314} & \textbf{0.5121} & \textbf{0.1766} & \textbf{0.1404} & \textbf{0.0589} & 95.41
  
  & \textbf{0.0624} & \textbf{0.2048} & \textbf{0.0655} & \textbf{0.00047} & \textbf{0.0110} & 2237.72
  \\
  \midrule
  \multirow{4}{*}{\textbf{FGCNN}} & Themis & 0.2662 & 0.5874 & 0.2067 & 0.1346 & 0.0650 & 181.09
  
  & 0.0707 & 0.2169 & 0.0683 & 0.00053 & 0.0114 & 16336.30
  \\
  \cdashline{3-14}[2pt/2pt]
  & TestSGD$_{\theta=0.01}$ & 0.0768 & 0.1771 & 0.0598  & 0.0473 & 0.0177 & 69.84
  
  & 0.0587 & 0.1404 & 0.0444  & 0.00012 & 0.0042 & 2893.24
  
  \\
  & TestSGD$_{\theta=0.005}$ & 0.1140 & 0.2533 & 0.0985  & 0.0936 & 0.0227 & 111.21
  
  & 0.0628 & 0.1641 & 0.0522  & 0.00018 & 0.0071 & 3088.18
  \\
    
  
  & \tabsysname & \textbf{0.2594} & \textbf{0.5523} & \textbf{0.2020} & \textbf{0.1225} & \textbf{0.0650} & 104.82
  
  & \textbf{0.0698} & \textbf{0.2059} & \textbf{0.0678} & \textbf{0.00053} &  \textbf{0.0114}  & 2297.14
  
  \\
  \midrule
    &  & \multicolumn{6}{c}{\textbf{BlackFriday}}  &  \multicolumn{6}{c}{\textbf{Amazon}} 
  \\
  \cmidrule(lr){3-8} \cmidrule(lr){9-14}
    &  & \multicolumn{1}{c}{$UF_{auc}$} & \multicolumn{1}{c}{$UF_{mrr}$} & \multicolumn{1}{c}{$UF_{ndcg}$} & \multicolumn{1}{c}{$UF_{urd}$} & \multicolumn{1}{c}{$UF_{urp}$} & \multicolumn{1}{c}{Time(s)} &
    \multicolumn{1}{c}{$UF_{auc}$}&
    \multicolumn{1}{c}{$UF_{mrr}$} & \multicolumn{1}{c}{$UF_{ndcg}$} & \multicolumn{1}{c}{$UF_{urd}$} & \multicolumn{1}{c}{$UF_{urp}$}  & \multicolumn{1}{c}{Time(s)}
  \\
  \midrule
  \multirow{4}{*}{\textbf{Wide\&Deep}} & Themis & 0.2473 & 0.8154 & 0.2730 & 0.3832 & 0.0703 & 173.92
  
  & 0.0757 & 0.1266 & 0.0453 & 0.0096 & 0.0656 & 1269.65
  \\
  \cdashline{3-14}[2pt/2pt]
  & TestSGD$_{\theta=0.005}$ & 0.0253 & 0.0296 & 0.0054  & 0.0430 & 0.0047 & 61.01 
  
  & 0.0372 & 0.0680 & 0.0239  & 0.0035 & 0.0547  & 811.84
  
  \\
  & TestSGD$_{\theta=0.002}$ & 0.2242 & 0.4728 & 0.1736  & 0.2216 & 0.0458 & 101.01
  
  & 0.0720 & \textbf{0.1266} & \textbf{0.0453} & 0.0034 & \textbf{0.0656}  & 1583.65
  \\
  
  
  & \tabsysname & \textbf{0.2472} & \textbf{0.7222 }& \textbf{0.2668} & \textbf{0.3424} & \textbf{0.0678} & 78.70
  
  & \textbf{0.0736} & 0.1164 & \textbf{0.0453} & \textbf{0.0081} & 0.0615 & 783.50
  \\
  \midrule
  \multirow{4}{*}{\textbf{DeepFM}} & Themis & 0.2823 & 0.7011 & 0.2409 & 0.3892 & 0.0591 & 201.43
  
  & 0.0767 & 0.1301 & 0.0453 & 0.0092 & 0.0652 & 1335.13
  \\
  \cdashline{3-14}[2pt/2pt]
   & TestSGD$_{\theta=0.005}$ & 0.0143 & 0.0855 & 0.0139  & 0.0838 & 0.0077 & 62.87
   
  & 0.0403 & 0.0632 & 0.0210  & 0.0071 & 0.0548  & 815.69
  \\
  
  & TestSGD$_{\theta=0.002}$ & 0.2245 & 0.4445 & 0.1581 & 0.2137 & 0.0414 & 120.62
  
  & \textbf{0.0718} & 0.1201 & \textbf{0.0453}  & 0.0076 & 0.0642  & 1657.17
  \\
    
   & \tabsysname & \textbf{0.2352} & \textbf{0.6667} & \textbf{0.2218} & \textbf{0.3667} & \textbf{0.0569} & 93.56
   
   & 0.0676 & \textbf{0.1235} & 0.0396  & \textbf{0.0082} & \textbf{0.0648}  & 720.26
   \\
  \midrule
  \multirow{4}{*}{\textbf{DCN}} & Themis & 0.2777 & 0.7500 & 0.2487 & 0.3830 & 0.0627 & 168.49
  
  & 0.0764 & 0.1219 & 0.0419 & 0.0078 & 0.0655 & 1374.06
  \\
  \cdashline{3-14}[2pt/2pt]
  & TestSGD$_{\theta=0.005}$ & 0.0105 & 0.0149 & 0.0094 & 0.0826 & 0.0063 & 62.44
  
  & 0.0375 & 0.0793 & 0.0271 & 0.0040 & 0.0547 & 854.64
  \\
  
  & TestSGD$_{\theta=0.002}$ & 0.2109 & 0.4503 & 0.1601  & 0.1000 & 0.0412  & 122.44
  
  & 0.0702 & \textbf{0.1215} & 0.0401 & 0.0062 & \textbf{0.0655} & 1666.68
  \\
    
  & \tabsysname & \textbf{0.2437} & \textbf{0.6511} & \textbf{0.2247} & \textbf{0.3181} & \textbf{0.0580} & 84.62
  
  & \textbf{0.0713} & 0.1157 & \textbf{0.0405} & \textbf{0.0076} & 0.0576 & 782.63
  \\
  \midrule
  \multirow{4}{*}{\textbf{FGCNN}} & Themis & 0.2441 & 0.8472 & 0.2865 & 0.4571 & 0.0725 & 192.88
  
  & 0.0714 & 0.1270  & 0.0431 & 0.0098 & 0.0656 & 1388.79
  \\
  \cdashline{3-14}[2pt/2pt]
  & TestSGD$_{\theta=0.005}$ & 0.0199 & 0.1034 & 0.0320  & 0.0621 & 0.0051 & 65.98
  
  & 0.0420 & 0.0695 & 0.0245 & 0.0040 & 0.0547 & 1010.55
  \\
  
  & TestSGD$_{\theta=0.002}$ & 0.1710 & 0.4719 & 0.1676 & 0.2045 & 0.0392 & 125.89
  
  & 0.0685 & \textbf{0.1270} & \textbf{0.0431} & 0.0078 & \textbf{0.0656} & 1708.27
  \\
    
 
  & \tabsysname & \textbf{0.2438} & \textbf{0.8371} & \textbf{0.2634} & \textbf{0.4539} & \textbf{0.0641} & 99.82
  
  & \textbf{0.0693} & 0.1219 & 0.0418 & \textbf{0.0094} & 0.0572 & 711.75
  \\
  \bottomrule
\end{tabular} }
\end{table*}

\textbf{Baselines.}
We are not aware of a fairness testing method specialized for DRSs from the perspective of multi-sensitive attributes.
Since Themis~\cite{galhotra2017fairness} and TestSGD~\cite{zhang2022testsgd} have explored the issues of multi-attributes group fairness in the classification tasks, we take them as the baselines by extending them to our tasks.
For fair comparison, all three methods group users based on the same sensitive attributes and quantify unfairness using the metrics introduced in Section~\ref{subsec:rs_evaluation_metrics}.
Specifically, in Themis, the target groups are found via brute-force enumeration based on sparse values of different attributes, which we consider as the globally optimal target groups, and the fairness scores it obtains represent the accurate values.
In TestSGD, a parameter $\theta$ is used to improve testing efficiency by filtering out groups with users less than $\theta$ percent of the total users. Then it calculates the fairness scores of the remaining groups and finds the target groups by traversal.
In our experiment, we set $\theta$ to 0.01, 0.005, and 0.002 for datasets of different scales to investigate the impact of $\theta$ on the testing results of TestSGD. 
Moreover, to mitigate the impact of long-tail distribution of super subgroups, we also filter out those groups with users less than 0.001 percent of the population for all the testing methods.

\textbf{Implementation.}
All experiments are conducted on a server running Ubuntu 1804 operating system with 1 Intel(R) Xeon(R) E5-2682 v4 CPU running at 2.50GHz, 64GB memory and 2 NVIDIA GeForce GTX 1080 Ti GPU. To mitigate the effect of randomness, all experiment results are the average of 5 runs.
The hyper-parameters used in our experiments are: a) inertial factor $\alpha=0.09$, b) the acceleration parameters $c_1=c_2=2$, and c) the velocity limits $v^*=2$,
which are obtained by experimental analysis and more details of the analysis will be discussed in the following section.

\subsection{Research Questions}
\label{subsec:rq}

\label{research_questions}
\textbf{\textit{RQ1: Is \sysname effective enough to reveal and measure the unfairness of a DRS?}} \label{q:1}

To answer this question, we evaluate \sysname on the four datasets and four DRSs using the five metrics described in Section~\ref{subsec:rs_evaluation_metrics}.
The main results are summarized in Table~\ref{tab:effectiveness_result}, in which each entry denotes the multi-attribute group fairness scores (Definition~\ref{def:multi-attributes_group_fairness}), which is the difference in the metric-performance values of the most advantaged and the disadvantaged groups as found by the corresponding testing method. The larger the score, the more unfair is the DRS.
The rows underlined with a dotted line denote the results of Themis, which are optimal values (corresponding to the optimal target groups) for each fairness metric.
\hz{In order to evaluate the effectiveness of TestSGD and \sysname, we define the \emph{testing accuracy} as the ratio of the entry of a testing method to that of Themis.}

Either TestSGD or \sysname may end up with the \emph{locally} optimal target groups.
The bold value in each column denotes the most accurate result, or the one closest to that of Themis.
Table~\ref{tab:effectiveness_result} presents us a variety of observations and insights as follows.
     
\emph{
First, \sysname almost always obtains the most accurate testing results (occupying the bold values on $\sim$90\% cases), i.e., achieving $\sim$95\% testing accuracy compared to 94\% and 43\% for TestSGD.}
A closer look reveals that for all the metrics, \sysname achieves an average of 95.27\%,97.37\%, 92.38\% and 92.83\% of the optimal results on MovieLens, LFM360K, BlackFriday and Amazon respectively.
Besides, the performance of \sysname is rather stable despite the random nature of DPSO.
For instance, even for LFM360k which has the most user groups, the test accuracy of \sysname only fluctuates in a small range of 88.68\%-100\% due to the specific optimization we designed for DRSs, as mentioned in Section~\ref{subsec:search_based_testing_with_dpso}, which greatly improves the stability of \sysname.

\begin{figure}[t]
    \setlength{\abovecaptionskip}{4pt}
    \centering
    \includegraphics[width=0.485\textwidth]{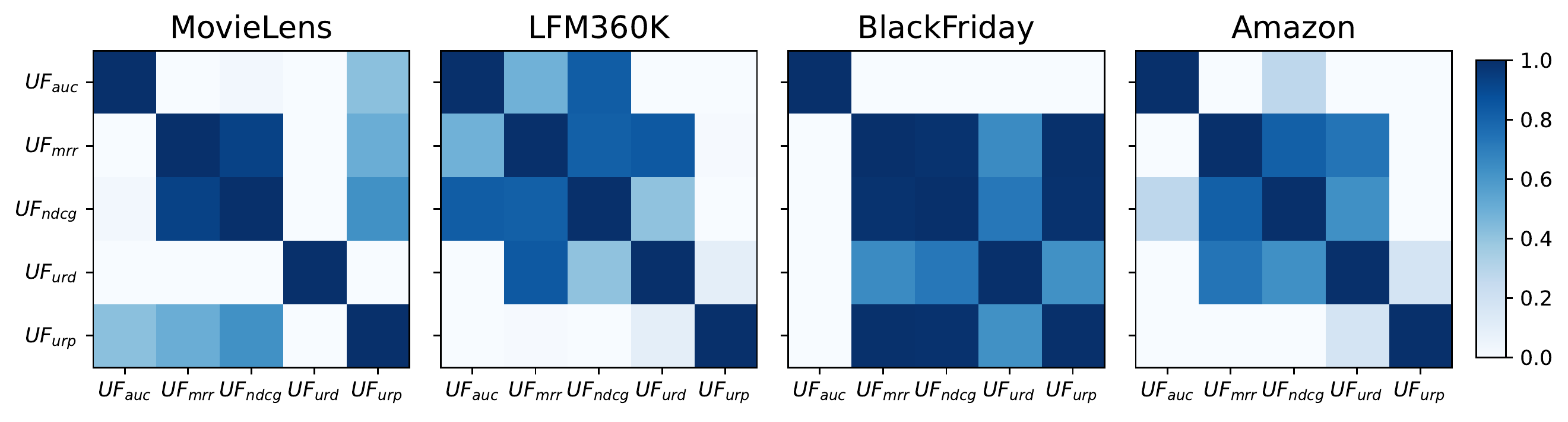}
    \caption{The correlation between different metrics.}
    \label{fig:correlation_heatmap}
\vspace{-0.60cm}
\end{figure}

The testing results can also be reliably used as evidence to assess the fairness deficiencies of different DRSs.
For example, considering MovieLens and $UF_{urd}$ (whether user gets diverse recommendations), with Themis we can conclude that Wide\&Deep is the most unfair with a difference of $0.1786$ (for FGCNN it is $0.1346$), indicating that Wide\&Deep presents users with the least diverse recommendations and echo chamber is easier to occur. 
With \sysname, we can get a similar conclusion that Wide\&Deep is more unfair than DCN and FGCNN.
However, TestSGD$_{\theta=0.01}$ would conclude incorrectly that Wide\&Deep is the most fair, deriving the smallest $UF_{urd}$ value. 
It is worth noticed that FGCNN exhibits the most serious fairness issues among the four DRSs in most cases. A reasonable explanation is that the feature intersection process of FGCNN is too complex, which cause it to be more sensitive to the individual information and tends to amplify individual differences.


In addition, we also designed a set of experiments to compare the testing accuracy of the three methods given a fixed testing budget, as shown in Table~\ref{tab:fixed_time_result}.
The testing time are dynamically set as follows,
\begin{equation}
\setlength{\abovedisplayskip}{2pt}
\setlength{\belowdisplayskip}{2pt}
\nonumber
    Time = 0.005 * (N_{users} + N_{groups}),
\end{equation}
where $N_{users}$ and $N_{groups}$ denote the number of users and groups after being divided based on sensitive attributes, respectively.

In Table~\ref{tab:fixed_time_result}, the largest value in each column is marked bold, representing the biggest difference in the target groups found in the given time. 
Note that different from Table~\ref{tab:effectiveness_result}, here the group difference found by Themis is not always the largest anymore, while that found by \sysname is for most cases (e.g., 79\% compared to 62\% of Themis and 55\% of TestSGD). 
The reason is that for the fixed-time testing, Themis degraded to random search which may not be able to find the target groups with the globally maximal difference, while \sysname is more effective than Themis and TestSGD in general. 
The results on BlackFriday and Amazon are attached in the Appendix.


\emph{Second, how much a DRS is fair is not purely determined by the model itself, but also influenced by the choice of the fairness metrics.} 
For instance, for BlackFriday, DeepFM achieves the smallest fairness score (being the most fair) regarding $UF_{urp}$ ($0.0591$ by Themis), while it also the most unfair regarding $UF_{auc}$, among the four DRSs.
What's more, we observe that fairness performance regarding different metrics may not be correlated in an expected way. 
For example, while all the three metrics $UF_{auc}, UF_{mrr}$ and $UF_{ndcg}$ describe how well the recommendations capture user interest, a DRS may not achieve the same levels of fairness for them.
Given MovieLens, DeepFM is the most fair regarding $UF_{auc}$, but is also the most unfair regarding $UF_{mrr}$.

To capture the observations above more in detail, we depict heat maps of correlation between different metrics with regard to DRSs, for the four datasets in Figure~\ref{fig:correlation_heatmap}.
Given a map, each block represents the correlation value between the two corresponding metrics (computed with their values taken for the four DRSs).
We can see that for all the four datasets, $UF_{mrr}$ is negatively correlated with $UF_{auc}$, while positively correlated with $UF_{ndcg}$.
This implies that if the replacement of the DRS improves $UF_{mrr}$, then it may increase (decrease) $UF_{ndcg}$ ($UF_{auc}$).
Also, in general the popularity metric $UF_{udp}$ seems to be negatively correlated with $UF_{auc}$, meaning if we want the system to know better whether users prefer popular items, then we may end up with the sacrificed performance.
More importantly, while it may usually be perceived that more diverse recommendations might reduce performance, we observe the opposite, namely $UF_{urd}$ and $UF_{auc}$ (or $UF_{mrr}$, $UF_{ndcg}$) are positively correlated sometimes.
Hence it is barely possible to achieve simultaneous improve on fairness of all the metrics, by choosing DRSs (none of the four DRSs is the most fair for all the metrics). 
And for a DRS developer, s/he has to consider the specific requirements of the application and carefully make trathe appropriate metrics.

\begin{figure}[t]
    \setlength{\abovecaptionskip}{1pt}
    \centering
    \includegraphics[width=0.335\textwidth]{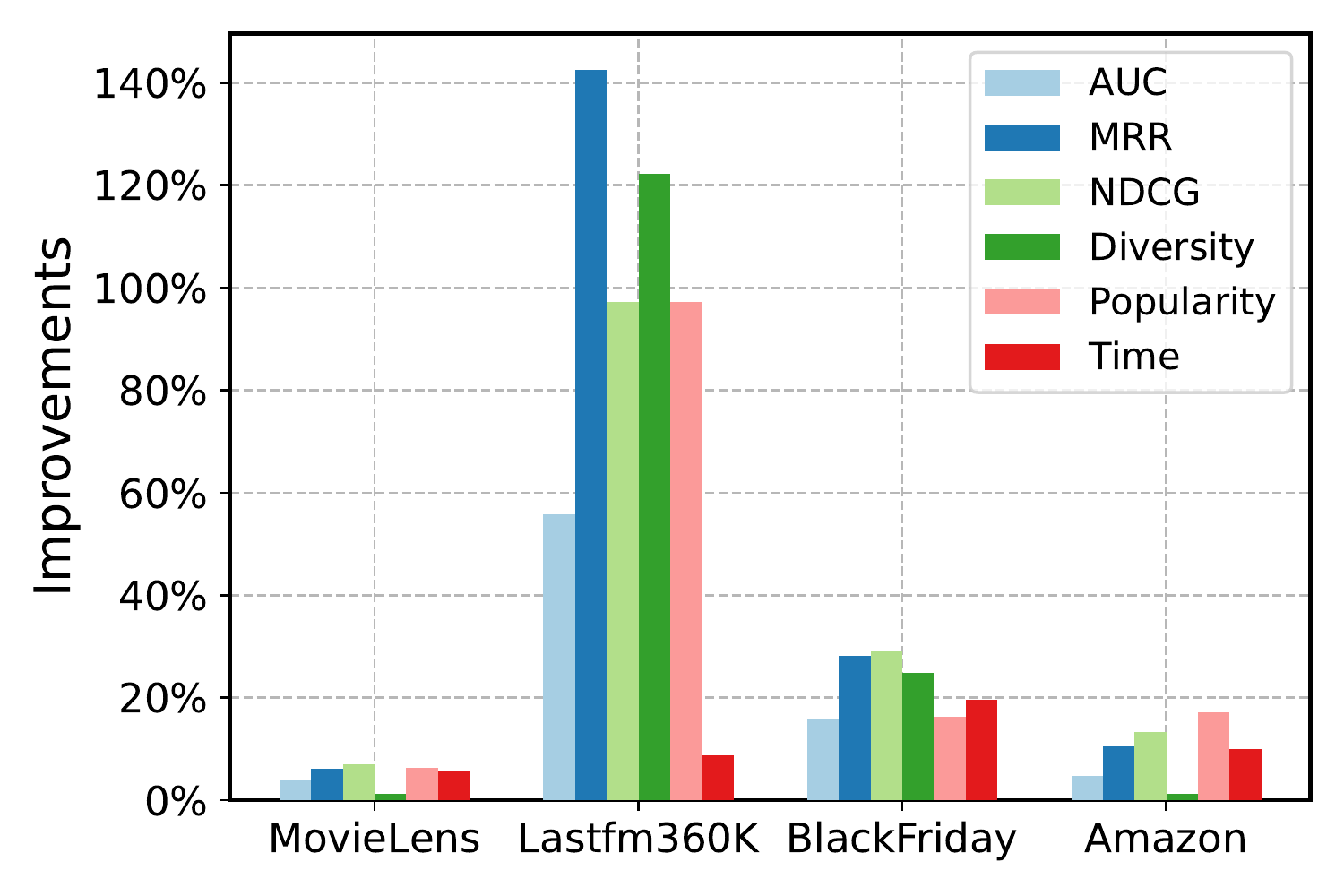}
    \caption{The effectiveness improvements of \sysname compared to the vanilla DPSO.}
    \label{fig:improvements_of_optimized_dpso}
   \vspace{-0.60cm}
\end{figure}

\emph{Third, we compare \sysname with the vanilla PSO to validate the delicate designs of our DPSO algorithm in \sysname.}
As shown in Figure~\ref{fig:improvements_of_optimized_dpso}, where the horizontal axis represents the different data sets, and the vertical axis represents the improvement obtained by the optimized method. 
On the four datasets, the test results of the optimized DPSO algorithm have improved by 4.90\%, 103.01\%, 22.3\% and 9.35\% respectively, while the average time consumption only increased by 11.01\%. 
In particular, in the case of sparse user distribution, the test results of the optimized algorithm improve by 103.01\% on average for LFM360K (the largest dataset), while the time consumption increases by only 8.75\%. 
The reasons behind is that our method makes the particles more concentrated in the densely distributed region of the user and then move quickly towards the optimal solution, thus automatically filtering many invalid groups and saving a lot of time for computation.

With all the observations above, we get the following answer:

\begin{center}
\begin{tcolorbox}[colback=gray!15,
                  colframe=black,
                  width=0.47\textwidth,
                  arc=1mm, auto outer arc,
                  boxrule=0.35pt,
                 ]
\textbf{Answer to RQ1:} \sysname is effective enough, even in limited time, to reveal and measure the unfairness of DRSs, achieving $\sim$95\% testing accuracy with $\sim$half to 1/8 time. Given fixed testing budget, \sysname is the most effective.
\end{tcolorbox}
\end{center}

\begin{table*}[tp]
\setlength{\abovecaptionskip}{6pt}
\setlength\tabcolsep{8pt}
\centering
\caption{The fairness testing results on MovieLens and LFM360K within limited time.}
\label{tab:fixed_time_result}
\resizebox{0.845\textwidth}{!}{
\begin{tabular}{*{14}{cccccccrcccccr}}
  \toprule
      &  & \multicolumn{5}{c}{\textbf{MovieLens}}  &  \multicolumn{5}{c}{\textbf{LFM360K}} 
  \\
  \cmidrule(lr){3-7} \cmidrule(lr){8-12}
    & & \multicolumn{1}{c}{$UF_{auc}$} & \multicolumn{1}{c}{$UF_{mrr}$} & \multicolumn{1}{c}{$UF_{ndcg}$} & \multicolumn{1}{c}{$UF_{urd}$} & \multicolumn{1}{c}{$UF_{urp}$} & 
    \multicolumn{1}{c}{$UF_{auc}$} &
    \multicolumn{1}{c}{$UF_{mrr}$} & \multicolumn{1}{c}{$UF_{ndcg}$} & \multicolumn{1}{c}{$UF_{urd}$} & \multicolumn{1}{c}{$UF_{urp}$}  
  \\
  \midrule
  \multirow{4}{*}{\textbf{Wide\&Deep}} & Themis & \textbf{0.2006} & 0.2158 & 0.1035 & 0.1114 & \textbf{0.0433} 
  & 0.0427 & 0.1528 & 0.0475 & 0.00025 & 0.0094  
  \\
  & TestSGD$_{\theta=0.01}$ & 0.0455 & 0.1357 & 0.0451  & 0.0379 & 0.0110  
  
  & 0.0360 & 0.1406 & 0.0443  & 0.00013 &  0.0011 
  \\
  & TestSGD$_{\theta=0.005}$ & 0.0655 & 0.1987 & 0.0673  & 0.0649  & 0.0274 
  
  & 0.0571 & 0.1715 & 0.0560  & 0.00024 &  0.0069 
  \\
    
   & \tabsysname & 0.1804 & \textbf{0.3478} & \textbf{0.1163} & \textbf{0.1202 }& 0.0390 
   
  & \textbf{0.0614} & \textbf{0.2010} & \textbf{0.0595} & \textbf{0.00041} & \textbf{0.0099}

  \\
  \midrule
  \multirow{4}{*}{\textbf{DeepFM}} & Themis & \textbf{0.1762 }& 0.2746 & 0.0889 & \textbf{0.1574} & 0.0220 
  & 0.0359 & 0.1495 & 0.0438 & \textbf{0.00051} & 0.0079 
  \\
   & TestSGD$_{\theta=0.01}$ & 0.0654 & 0.1547 & 0.0506  & 0.0542 & 0.0117 
   
  & 0.0376 & 0.1383 & 0.0433  & 0.00010 & 0.0025
  \\
  & TestSGD$_{\theta=0.005}$ & 0.0849 & 0.2251 & 0.0679  & 0.0728 & 0.0220 
  
  & 0.0598 & 0.1700 & 0.0547 & 0.00031 & 0.0069 
  \\ 
    
   & \tabsysname & 0.1661 & \textbf{0.3965} & \textbf{0.1214} & 0.1217 & \textbf{0.0272} 
   
  & \textbf{0.0603} & \textbf{ 0.2064} & \textbf{0.0610} & 0.00049 & \textbf{0.0102}
  \\
  \midrule
  \multirow{4}{*}{\textbf{DCN}} & Themis & 0.1315 & 0.3035 & 0.1400 & \textbf{0.0949} & 0.0298 
  
  & 0.0390 & 0.1446 & 0.0445 & 0.00025 & 0.0080 

  \\
  & TestSGD$_{\theta=0.01}$ & 0.0515 & 0.1273 & 0.0483 & 0.0402 & 0.0099  
  
  &  0.0379 & 0.1410 & 0.0444 & 0.00010 & 0.0017 
  \\
  
  & TestSGD$_{\theta=0.005}$ & 0.0803 & 0.1933 & 0.0676  & 0.0794 & 0.0298 
  
  & 0.0586 & 0.1687 & 0.0537 & 0.00025 & 0.0069 
  \\
    
  & \tabsysname & \textbf{0.2136} & \textbf{0.3086} & \textbf{0.1660} & 0.0912 & \textbf{0.0304} 
  
  & \textbf{0.0610} & \textbf{0.2083} & \textbf{0.0635} & \textbf{0.00047} & \textbf{0.0104}
  \\
  \midrule
  \multirow{4}{*}{\textbf{FGCNN}} & Themis & 0.0964 & 0.2370 & 0.0855 & 0.1086 & \textbf{0.0488}  
  
  & 0.0349 & 0.1487 & 0.0461 & 0.00025 & 0.0079 
  \\
  
  & TestSGD$_{\theta=0.01}$ & 0.0768 & 0.1652 & 0.0586  & 0.0375 & 0.0177 
  
  & 0.0397 & 0.1404 & 0.0443  & 0.00011 & 0.0022 
  
  \\
  & TestSGD$_{\theta=0.005}$ & 0.0952 & 0.2147 & 0.0801  & 0.0935 & 0.0230 
  
  & 0.0586 & 0.1641 & 0.0521  & 0.00018 & 0.0068 
  \\
    
  
  & \tabsysname & \textbf{0.2379} &\textbf{ 0.3522} & \textbf{0.1198} & \textbf{0.1103} & 0.0274 
  
  & \textbf{0.0661} & \textbf{0.1973} & \textbf{0.0622} & \textbf{0.00046} &  \textbf{0.0108}
  
  \\
  \bottomrule
\end{tabular} }
\end{table*}

\noindent \textbf{\textit{RQ2: How efficient is \sysname compared with existing work?}} 

To answer this question, we evaluate the efficiency of \sysname and make comparisons with Themis and TestSGD.
The main quantitative comparison results are shown in Table~\ref{tab:effectiveness_result}.
It is clearly observed that \sysname can achieve comparable testing performance to Themis ($\sim$95\% testing accuracy) in around $\sim$half to 1/8 of the time Themis requires.
Moreover, the advantage of \sysname in efficiency becomes more significant when there are more users and sensitive attributes in the testing candidates.
For instance, on the LFM360K dataset, \sysname has promoted the testing efficiency by more than 5 times and 1 time compared to Themis and TestSGD$_{\theta=0.002}$, respectively. 
Specifically, Themis requires more than $4.5$ hours for testing just $260,000$ users, which is almost infeasible in the industry scenarios with critical efficiency needs.

\begin{figure}[t]
    \setlength{\abovecaptionskip}{6pt}   
    \centering
    \includegraphics[width=0.43\textwidth]{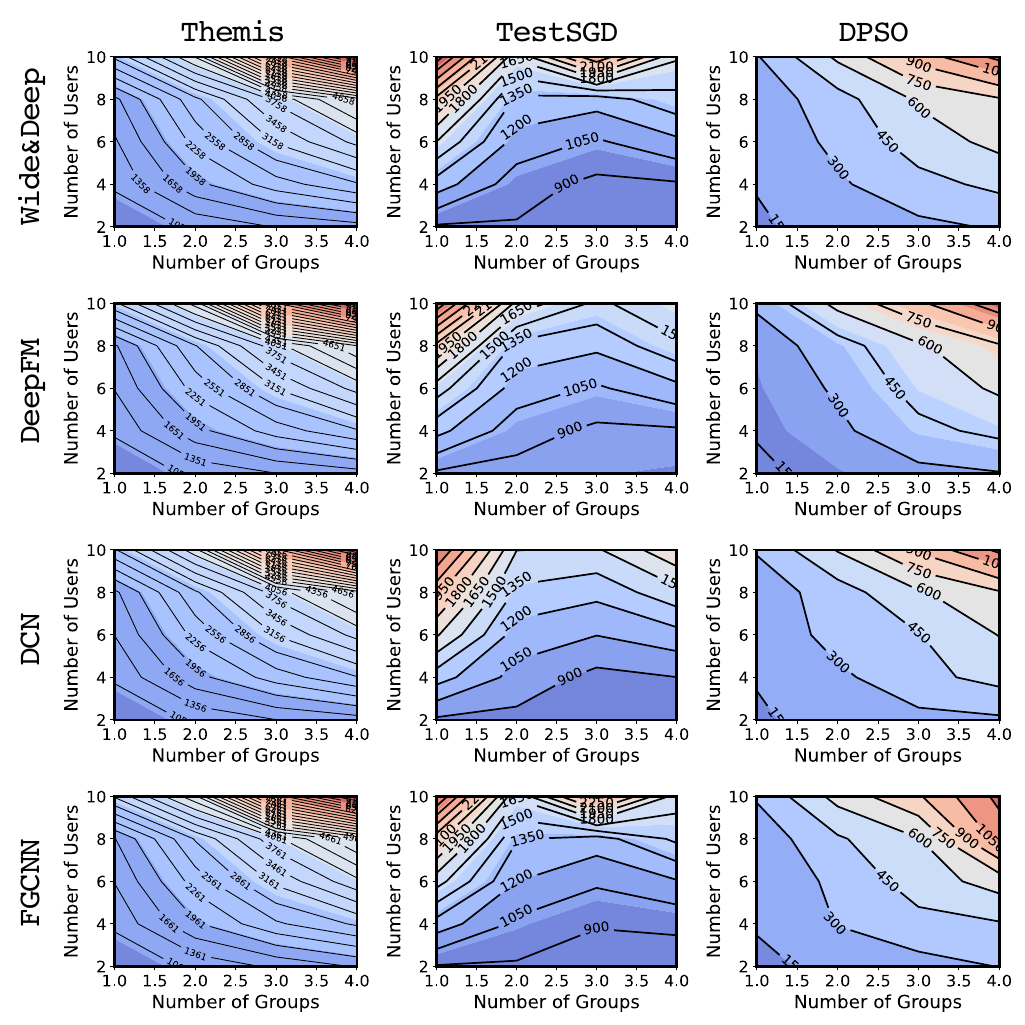}
            \caption{Comparison of the efficiency of the three methods.
            The horizontal axis indicates the number of groups after division and vertical axis represents the total number of users, with unit of ten thousand.
            The contour lines in the graph indicate the time required for the test. 
            Intuitively, denser contours indicate a higher slope of time growth.
            The different colored areas in the graph indicate the distribution of the values of the test time, with blue areas indicating smaller values and red indicating more significant time consumption.}
    \label{fig:efficiency_comparision}
    \vspace{-0.45cm}
\end{figure}

Considering that numbers of users and groups are two critical factors affecting the testing efficiency, in this experiment, we further explore their impact on the testing efficiency of the three testing approaches.
The results are shown in Figure~\ref{fig:efficiency_comparision}, in which the horizontal axis denotes the number of groups after division and vertical axis represents the total number of users. From Figure~\ref{fig:efficiency_comparision}, we can see that the time consumption of Themis increases exponentially as the number of groups and users increases.
Concretely, in the test against DCN, as the number of users increased from 20,000 to 100,000 and the number of groups increased from 10,000 to 40,000, the time consumption increased from 756 seconds to 9768 seconds.
Such level of efficiency is unacceptable to the realistic industry recommendation scenarios, where there are usually tens of millions or even hundreds of millions of users.
It also shows a similar trend on the efficiency of TestSGD as the number of users grows.
When the total number of users remains the same and the number of groups increases, i.e., the user distribution becomes more sparse, TestSGD thus filters out more groups that do not meet the number requirement, so its testing efficiency becomes higher.
However, it comes at the expense of testing effectiveness and the improvement is not significant enough either, which is also corroborated in Table~\ref{tab:effectiveness_result}.

In comparison, the growth trend on time consumption of \sysname is relatively smoother when the number of users and groups increases.
Concretely, the time overhead of \sysname increases from 114 seconds to 1242 seconds as the test scenario changes from the simple to the most complex, which are only $15\%$ and $12\%$ of the time required by Themis under the same conditions, respectively.
We believe such high efficiency mainly benefits from the three delicate designs in our \sysname.
First, the distribution-based particle initialization modeled from testing candidates can effectively improve the coverage of particles, thus helping to quickly locate the target area in the search space.
Second, the specifically designed double-ended search scheme can boost the search process and guarantee the search convergence.
Besides, the designed \infobase can help different particle swarms share information with each other, which would avoid many unnecessary computational overheads.

\begin{center}
\begin{tcolorbox}[colback=gray!15,
                  colframe=black,
                  width=0.47\textwidth,
                  arc=1mm, auto outer arc,
                  boxrule=0.35pt,
                 ]
\textbf{Answer to RQ2:} 
\sysname's testing efficiency outperforms Themis and TestSGD in a significant margin and is more efficient in more complex test scenarios.
\end{tcolorbox}
\end{center}

\begin{figure}
    \setlength{\abovecaptionskip}{1pt}
    \centering
    \subfigure[Impact of particle size]{
        \centering
        \includegraphics[width=0.22\textwidth]{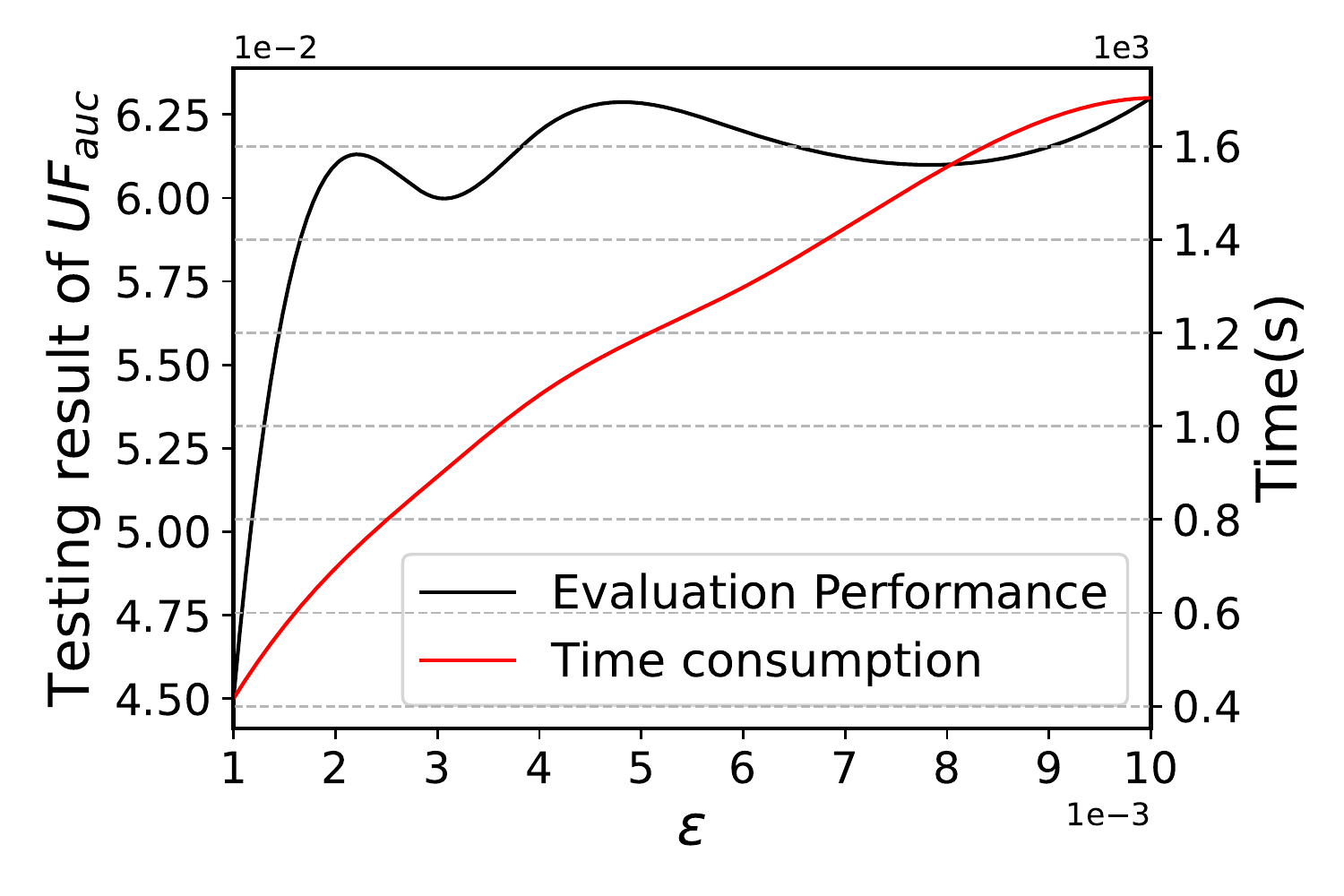}
        \label{fig:size_experiment}
    }
    \subfigure[Impact of iterations]{
        \centering
        \includegraphics[width=0.22\textwidth]{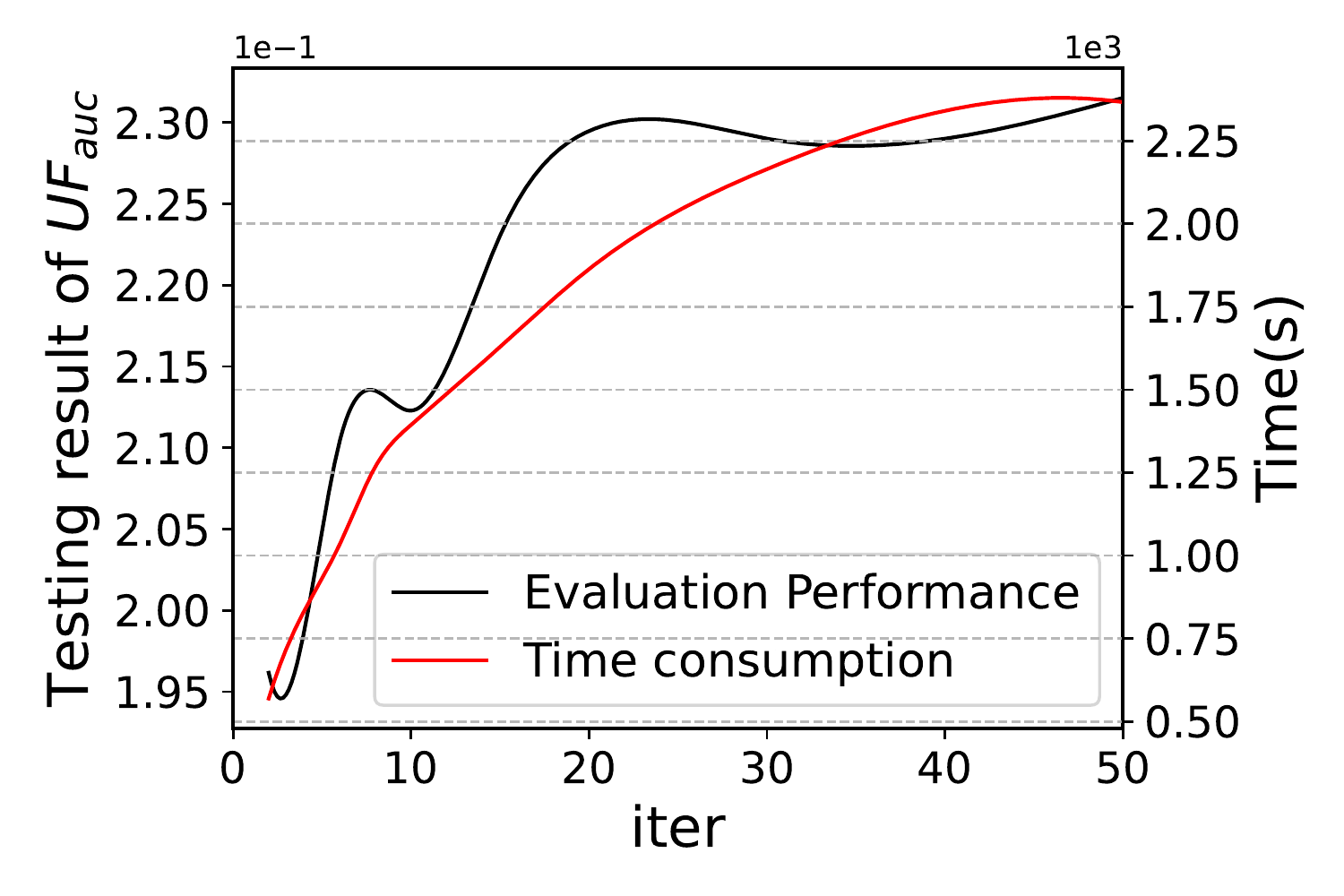}
      \label{fig:iter_experiment}
    }
    \caption{The impact of $\epsilon$ and $iter$ on efficacy of \sysname.} 
    \label{fig:hyperparameter_experiment}
    \vspace{-0.25cm}
\end{figure}

\textbf{Hyperparametric Experiment.} Usually, the size of the initialized populations and the maximum number of iterations influence the accuracy and efficiency of the testing.
We design the hyperparametric experiment to analyze the effect of these two hyperparameters. 
Define $\epsilon = n_{pt} / |N_{groups}|$, which indicates the ratio of the initialized particles to the set of user groups.
The result is shown in Figure~\ref{fig:size_experiment}.
The smaller $\epsilon$ is, the more difficult it is for the particles to collect enough information in a high-dimensional search space, resulting in low accuracy.
As $\epsilon$ increases, the accuracy of \sysname gets more stable, and the testing result is getting closer to the optimal result while the testing time becomes longer.
We suggest choosing $\epsilon$ in the range of [0.002, 0.005] in the complex case, which can ensure both high testing accuracy and testing efficiency.
Then we set $\epsilon = 0.005$ to discuss the impact of the number of iterations as shown in Figure~\ref{fig:iter_experiment}. 
As $n_{it}$ increases, the overall testing accuracy improves and eventually approaches the optimum.
In this case, we suggest choosing $n_{it}$ in the range of [15, 20].
Note that in the relatively simple cases such as MovieLens and Amazon where $N_{groups} \leq 1,000$, we set the $\epsilon$ in the range of [0.1, 0.2] and $n_{it}$ in the range of [5,10] to get better results.
Our results on the two hyperparameters provide testers hindsight of how to make a trade-off between testing efficiency and accuracy. 



\noindent \textbf{\textit{RQ3: Can we use the the testing results of \sysname to improve the fairness of DRSs?}} 

\begin{table}[t]
\setlength{\abovecaptionskip}{6pt}
\setlength\tabcolsep{2pt}
\renewcommand\arraystretch{1.15}
\centering
\caption{Fairness mitigation based on our testing results.}
\label{tab:mitigation_result_v2}
\resizebox{0.475\textwidth}{!}{
\begin{tabular}{*{10}{cccccccccc}}
  \toprule
      &  & \multicolumn{4}{c}{\textbf{MovieLens}}  &  \multicolumn{4}{c}{\textbf{Amazon}} 
  \\
  \cmidrule(lr){3-6} \cmidrule(lr){7-10}
    & &  \multicolumn{1}{c}{$MRR \uparrow$} & \multicolumn{1}{c}{$NDCG \uparrow$} & \multicolumn{1}{c}{$URD \uparrow$} & \multicolumn{1}{c}{$URP\downarrow$} &   \multicolumn{1}{c}{$MRR \uparrow$} & \multicolumn{1}{c}{$NDCG \uparrow$} & \multicolumn{1}{c}{$URD \uparrow$} & \multicolumn{1}{c}{$URP \downarrow$}
  \\
  \midrule
  \multirow{2}{*}{\textbf{Wide\&Deep}} & Original & 0.1786 & 0.0728 & 0.7625 & 0.0122
  & 0.3041 & 0.1163 & 0.9884 & 0.0241 
  \\

  & Mitigated & \textbf{0.3208} & \textbf{0.1226} &\textbf{ 0.7967} & \textbf{0.0113 }
   
  & \textbf{0.3909} & \textbf{0.1489} & \textbf{0.9905} & \textbf{0.0169}

  \\
  \midrule
  \multirow{2}{*}{\textbf{DeepFM}} & Original  & 0.1783 & 0.0787 & 0.7771 & 0.0119
  & 0.3076 & 0.1197 & 0.9879 & 0.0245
  \\
  & Mitigated & \textbf{0.3352} & \textbf{0.1369} & \textbf{0.8089} & \textbf{0.0102}
  
  & \textbf{0.4181} & \textbf{0.1512} & \textbf{0.9901} & \textbf{0.0172}
  \\
  \midrule
  \multirow{2}{*}{\textbf{DCN}} & Original & 0.1824 & 0.0743 & 0.7705 & 0.0113  
  
  & 0.3027 & 0.1174 & 0.9882 & 0.0227
  \\
  & Mitigated & \textbf{0.3480} & \textbf{0.1292} & \textbf{0.8150} & \textbf{0.0106}
  
  & \textbf{0.4122} & \textbf{0.1508} & \textbf{0.9906} & \textbf{0.0169}
  \\
  \midrule
  \multirow{2}{*}{\textbf{FGCNN}} & Original & 0.1773 & 0.0796 & 0.7650 & 0.0128
  
  & 0.2985 & 0.1125 & 0.9886 & 0.0245
  \\
  
  & Mitigated & \textbf{0.3413} & \textbf{0.1395} & \textbf{0.8088} & \textbf{0.0114}
  
  & \textbf{0.4003} & \textbf{ 0.1467} & \textbf{0.9906} & \textbf{ 0.0142}
  \\
  \bottomrule
\end{tabular} }
\end{table}

Besides \emph{uncovering and evaluating} the severity of fairness problem, our testing also aims to provide insight and guidance on \emph{improving} the group fairness of a DRS.
Note that our testing report not only includes the multi-attribute group fairness scores as shown in Table~\ref{tab:effectiveness_result}, but also provides a user-defined number of advantaged groups and disadvantaged groups (remember that the \infobase records the metric values of each group).
In this way, a model developer can target any specific disadvantaged groups (not necessarily the worst one) regarding a specific metric for bias mitigation.

We adopt a simple re-ranking based method~\cite{gomez2021winner} in the experiment to improve the recommendation performance over disadvantaged groups to show the usefulness of fixing these groups in improving the overall fairness.
We selected the 10\% groups with the worst performance in $MRR$, $NDCG$, $ URD $, and $ URP $ for optimization, respectively.
To better demonstrate the effect of bias mitigation, we first directly show the changes in the relevant metrics for these groups before and after optimization, as shown in Table~\ref{tab:mitigation_result_v2}.
For example, considering MovieLens and Wide\&Deep, $MRR$ increased by 87.72\%, meaning that disadvantaged groups have more access to their target items.
In addition, considering Amazon and DeepFM, $URP$ decreases by 29.79\%, which indicates that the recommendation results are more in line with popularity preferences of disadvantaged groups. The change of overall fairness score is shown in Table \ref{tab:mitigation_result} in the Appendix due to space limit.

\begin{center}
\begin{tcolorbox}[colback=gray!15,
                  colframe=black,
                  width=0.47\textwidth,
                  arc=1mm, auto outer arc,
                  boxrule=0.35pt,
                 ]
\textbf{Answer to RQ3:} The test results of \sysname can provide valuable insight and guidance for bias mitigation.
\end{tcolorbox}
\end{center}

\section{Related work}
\label{sec:re}

\textbf{Fairness testing.} 
This work is closely related to fairness testing of deep learning models. 
There are many existing works focusing on individual fairness testing~\cite{fan2022explanation,zhang2020white}. 
They try to generate test cases that are discriminated on sensitive attributes, e.g., by changing the value of a sensitive attribute. 
In terms of group fairness, a number of definitions have been proposed~\cite{mitchell2021algorithmic,castelnovo2022clarification}.
\hz{In regards to DRSs, our main objective is to identify and address any discrimination that may be present towards real users within the recommendation system, rather than attempting to expose all instances of discrimination across the population. 
Therefore, while individual fairness testing methods based on generated inputs is not applicable, we focus on group fairness testing in this work.
However, the current level of attention towards group fairness testing is deemed insufficient in contemporary scholarship.}
Si et al.~\cite{si2021testing} proposed a statistical testing framework to detect whether a model satisfy multiple notions of group fairness, such as equal opportunity, equalized odds, etc.
Galhotra et al. proposed Themis~\cite{angell2018themis,galhotra2017fairness} to measure software group discrimination with two metrics, i.e., group discrimination score and causal discrimination score. 
Themis groups users by various sensitive attribute values and then gets the group fairness score for each group by brute-force enumeration.
Zhang et al. proposed TestSGD~\cite{zhang2022testsgd} to measure group discrimination. 
It automatically generates an interpretable rule set (i.e., reference for grouping) and each rule can be used to dichotomize the population. 
TestSGD filters the groups with a size smaller than a certain threshold like $0.01$ and then takes brute-force enumeration through the remaining groups to obtain their fairness scores.
\hz{However, the testing methods based on group fairness mentioned above are mostly focused on software and classification systems, which are intrinsically different to DRSs.}

\textbf{Fairness of recommender systems.}
\hz{The fairness of recommender systems is nowadays receiving more and more attention~\cite{rahmani2022experiments,beutel2019fairness}.
Recommender systems involve multiple stakeholders, including users, content/product providers, and side stakeholders~\cite{abdollahpouri2019multi}. 
There has been some work discussing fairness from user perspective~\cite{islam2021debiasing,leonhardt2018user}.
Yao et al. in~\cite{yao2017beyond} proposed four group fairness metrics to evaluate collaborative filtering models. 
They dichotomize the population by gender and study sexism in recommender systems on the MovieLens dataset as well as on synthetic datasets.
Li et al.~\cite{li2021user} divided users into active and inactive groups according to their behavioral history in the e-commerce recommender system. 
They find that users less active were treated significantly worse than those who are more active.
Wu et al.~\cite{wu2021fairness} introduced a sensitive attribute predictor to measure the association between news recommendations and gender, and used it to evaluate the fairness of the news recommender system.
However, the above mentioned works all use a single attribute to evaluate the fairness of recommender systems, which cannot capture the deeply hidden fairness issues in a multi-attribute perspective (Refer to Table~\ref{tab:multi_attribute} in Appendix).
In addition, they do not propose efficient methods to test the group fairness for DRSs.}

\textbf{Unfairness mitigation.}
\hz{Methods for unfairness mitigation can be divided into three main categories: pre-processing, in-processing and post-processing~\cite{chen2023bias}.
Pre-processing methods ensure fairness by eliminating the bias present in the training data.
Ryosuke et al.~\cite{sonoda2021pre} proposed a pre-processing method based on pairwise ordering for weighting pairs of training data, which can improve the fairness of the ranking model.
In-processing methods consider fairness constraints in the model training process.
For example, Yao et al. in~\cite{yao2017beyond} proposed four unfairness metrics, which can be optimized by adding fairness terms to the learning objective. 
Wu et al.~\cite{wu2021fairness} introduced an unfairness mitigation approach for news recommendation using decomposed adversarial learning, which eliminated the biases brought by sensitive features to ensure users get unbiased recommendations.
Post-processing method adjusts the output of the base recommendation model to mitigate unfairness.
Li et al.~\cite{li2021user} employed a re-ranking algorithm to produce new recommendation lists for users by leveraging the initial recommendation results, thereby meeting the fairness constraint.
Compared with the previous two methods for mitigating unfairness, post-processing methods do not require retraining the existing recommendation models, saving a significant amount of resources, and are also more practically feasible in real-world systems.
In this work, we employ post-processing methods to demonstrate how the results of \sysname can help disadvantaged groups and alleviate fairness concerns across the entire DRS.
}

\section{Conclusion}
\label{sec:con}
In this work, we study the problem of multi-attribute group fairness testing on deep learning based recommender systems by presenting \sysname, a systematical fairness testing framework built upon a specifically designed search-based testing algorithm.
We answer the three key research questions through quantitative and qualitative experiments on four extensively used recommendation models over four public benchmarks.
The experiment results demonstrate that \sysname is effective and efficient in revealing the fairness problems of DRSs from different perspectives, and outperforms the compared baselines by a significant margin.
Furthermore, it also shows that our testing results can provide insight and guidance for mitigating the bias of the tested DRSs with little negative impact on the recommendation performance.
Our work may shed new light on the research of building more fair deep recommender systems from a testing point of view and benchmark future testing research in this area.
\hz{Furthermore, \sysname has the potential to be used as a generic testing framework in other systems for multi-attribute group fairness testing.}

\section*{ACKNOWLEDGEMENT}
This research was supported by the Key R\&D Program of Zhejiang (2022C01018), the NSFC Programs (62102359 and 62106223), Alibaba Group through Alibaba Innovative Research Program, and the Research Center of Alibaba Artificial Intelligence Governance.
In addition, We are very grateful to the anonymous reviewers for their valuable comments to improve our paper. 


\bibliographystyle{ACM-Reference-Format}
\bibliography{ref} 

\clearpage
\section*{Appendix}

Table~\ref{tab:multi_attribute} shows the results of the fairness testing after dividing the population based on combinations of different sensitive attributes, which allows us to investigate the fairness issues under a multi-attribute view.
We select the dataset MovieLens (with sensitive attributes \emph{gender, age and occupation}) and BlackFriday (with sensitive attributes \emph{gender,age, occupation, city category, years of stay in the current city, and marital status }).
We can observe that with more sensitive attributes considered, more serious fairness issues can be disclosed.
For example, for MovieLens, when only Gender is considered (i.e., there are only two groups), the fairness-performance gap between the most advantaged group and disadvantaged group is $0.0021$.
However, when Gender, Age and Occupation are all considered (i.e., users get divided w.r.t the combinatorial values of the three dimensions), that gap turns to be $0.2537$, implying the recommendations are more unfair.

\begin{table}[!h]
\vspace{-0.95cm}
\setlength{\abovecaptionskip}{6pt}
\setlength\tabcolsep{4pt}
\setlength{\belowcaptionskip}{2pt}
\centering
\caption{Multi-attribute fairness based on different cases.}
\label{tab:multi_attribute}
\resizebox{0.475\textwidth}{!}{
\begin{tabular}{*{7}{ccccccc}}
  \toprule
   \textbf{Datasets} & \textbf{Attributes} & $UF_{auc}$ &$UF_{mrr}$ & $UF_{ndcg}$ & $UF_{div}$ & $UF_{pop}$ \\
  \midrule
   \multirow{3}{*}{\textbf{MovieLens}} &Gender & 0.0021 & 0.0163 & 0.0063 & 0.0014 & 0.0034\\
  \cmidrule(lr){2-7}
   &Gender,Age & 0.0324 & 0.0998 & 0.0332  & 0.0237 & 0.0137 \\
  \cmidrule(lr){2-7}
  &Gender,Age,Occupation & 0.2537 & 0.4819 & 0.1719  & 0.1786 & 0.0601
  \\
  \midrule
  \multirow{6}{*}{\textbf{BlackFriday}} &$A_1$ & 0.0096 & 0.0345 & 0.0134 & 0.0112 & 0.0027
  \\
  \cmidrule(lr){2-7}
   &\textbf{$A_2$} & 0.0456 & 0.1086 & 0.0406  & 0.0396 & 0.0094\\
  \cmidrule(lr){2-7}
  &\textbf{$A_3$} & 0.0670 & 0.1864 & 0.0695  & 0.0985 & 0.0187\\
  \cmidrule(lr){2-7}
   &\textbf{$A_4$} & 0.2019 & 0.6136 & 0.2096  & 0.2285 & 0.0247\\
  \cmidrule(lr){2-7}
  &\textbf{$A_5$} & 0.2969 & 0.6611 & 0.2430  & 0.3111 & 0.0681 \\
  \cmidrule(lr){2-7}
  &\textbf{$A_6$} & 0.2473 & 0.8154 & 0.2730 & 0.3832 & 0.0703\\
  \bottomrule
\end{tabular}}
\vspace{-0.95cm}
\end{table}

Note: 
\begin{itemize}
    \item $A_1$: Gender. \\
    
    \item $A_2$: Gender, Age. \\
    
    \item $A_3$: Gender, Age, City category. \\
    
    \item $A_4$: Gender, Age, City category, Years of stay in the current city. \\
    
    \item $A_5$: Gender, Age, City category, Years of stay in the current city,\\ Marital status. 
    
    \item $A_6$: Gender, Age, City category, Years of stay in the current city, Marital status, Occupation.  
\end{itemize}

Table~\ref{tab:mitigation_result} shows the change of overall fairness scores before and after mitigation mentioned in the RQ3 of section~\ref{subsec:rq}.
Each entry represents the gap between the most advantaged and disadvantaged groups is reduced w.r.t the corresponding metric after mitigation is applied.
We can notice that after mitigation, the fairness-performance w.r.t each metric gets improved (i.e., reduced gap between the target groups).

\begin{table}[H]
\setlength\tabcolsep{2pt}
\renewcommand\arraystretch{1.15}
\centering
\caption{Unfairness mitigation based on our testing results.}
\label{tab:mitigation_result}
\resizebox{0.475\textwidth}{!}{
\begin{tabular}{*{10}{cccccccccc}}
  \toprule
      &  & \multicolumn{4}{c}{\textbf{MovieLens}}  &  \multicolumn{4}{c}{\textbf{Amazon}} 
  \\
  \cmidrule(lr){3-6} \cmidrule(lr){7-10}
    & &  \multicolumn{1}{c}{$UF_{mrr}$} & \multicolumn{1}{c}{$UF_{ndcg}$} & \multicolumn{1}{c}{$UF_{urd}$} & \multicolumn{1}{c}{$UF_{urp}$} &   \multicolumn{1}{c}{$UF_{mrr}$} & \multicolumn{1}{c}{$UF_{ndcg}$} & \multicolumn{1}{c}{$UF_{urd}$} & \multicolumn{1}{c}{$UF_{urp}$} 
  \\
  \midrule
  \multirow{2}{*}{\textbf{Wide\&Deep}} & Original & 0.4603 & 0.1622 & 0.1605 & 0.0592
  & 0.1164 & 0.0453 & 0.0081 & 0.0615 
  \\

  & Mitigated & 0.3805 & 0.1270 & 0.1082 & 0.0534 
   
  & 0.0896 & 0.0355 & 0.0075 & 0.0572

  \\
  \midrule
  \multirow{2}{*}{\textbf{DeepFM}} & Original  & 0.5772 & 0.2009 & 0.1646 & 0.0594
  & 0.1235 & 0.0396  & 0.0082 & 0.0648
  \\
  & Mitigated & 0.3959 & 0.1309 & 0.1164 & 0.0525
  
  & 0.0927 & 0.0331 & 0.0072 & 0.0584
  \\
  \midrule
  \multirow{2}{*}{\textbf{DCN}} & Original & 0.5121 & 0.1766 & 0.1404 & 0.0589  
  
  & 0.1157 & 0.0405 & 0.0076 & 0.0576
  \\
  & Mitigated & 0.3825 & 0.1269 & 0.1044 & 0.0526
  
  & 0.0953 & 0.0346 & 0.0070 & 0.0521
  \\
  \midrule
  \multirow{2}{*}{\textbf{FGCNN}} & Original & 0.5523 & 0.2020 & 0.1225 & 0.0650
  
  & 0.1219 & 0.0418 & 0.0094 & 0.0572
  \\
  
  & Mitigated & 0.3954 & 0.1241 & 0.1031 & 0.0549
  
  & 0.1054 & 0.0365 & 0.0082 &  0.0536
  \\
  \bottomrule
\end{tabular} }
\end{table}

Table~\ref{tab:fixed_time_result_appendix} in the next page shows the fixed-time testing results on BlackFriday and Amazon as mentioned in the RQ1 of section~\ref{subsec:rq}.
We can observe that in most cases, our \sysname finds the largest gap between the advantaged and disadvantaged groups, revealing the severity of the fairness problem more accurately.

\begin{table*}[t]
\setlength\tabcolsep{8pt}
\centering
\caption{The fairness testing results on BlackFriday and Amazon within limited time.}
\label{tab:fixed_time_result_appendix}
\resizebox{0.85\textwidth}{!}{
\begin{tabular}{*{14}{cccccccrcccccr}}
  \toprule
    &  & \multicolumn{5}{c}{\textbf{BlackFriday}}  &  \multicolumn{5}{c}{\textbf{Amazon}} 
  \\
  \cmidrule(lr){3-7} \cmidrule(lr){8-12}
    &  & \multicolumn{1}{c}{$UF_{auc}$} & \multicolumn{1}{c}{$UF_{mrr}$} & \multicolumn{1}{c}{$UF_{ndcg}$} & \multicolumn{1}{c}{$UF_{urd}$} & \multicolumn{1}{c}{$UF_{urp}$} &
    \multicolumn{1}{c}{$UF_{auc}$}&
    \multicolumn{1}{c}{$UF_{mrr}$} & \multicolumn{1}{c}{$UF_{ndcg}$} & \multicolumn{1}{c}{$UF_{urd}$} & \multicolumn{1}{c}{$UF_{urp}$}  
  \\
  \midrule
  \multirow{4}{*}{\textbf{Wide\&Deep}} & Themis & 0.2263 & \textbf{0.7182} & 0.2267 & 0.1916 & 0.0545  
  
  & 0.0641 & 0.0836 & 0.0308 & 0.0084 & \textbf{0.0656} 
  \\
  & TestSGD$_{\theta=0.005}$ & 0.0253 & 0.0296 & 0.0054  & 0.0430 & 0.0047  
  
  & 0.0351 & 0.0572 & 0.0215  & 0.0029 & 0.0547   
  
  \\
  & TestSGD$_{\theta=0.002}$ & 0.2242 & 0.4702 & 0.1691  & 0.1765 & 0.0323  
  
  & 0.0426 & 0.0598 & 0.0220 & 0.0055 & \textbf{0.0656}  
  \\
  
  
  & \tabsysname & \textbf{0.2297} & 0.6818 & \textbf{0.2509} &\textbf{ 0.2809} & \textbf{0.0661}  
  
  & \textbf{0.0721} & \textbf{0.1066} & \textbf{0.0402} & \textbf{0.0076} & 0.0634  
  \\
  \midrule
  \multirow{4}{*}{\textbf{DeepFM}} & Themis & 0.1842 & 0.4464 & 0.1344 & 0.2238 & 0.0550 
  
  & 0.0647 & 0.1000 & 0.0342 & \textbf{0.0083} & 0.0641 
  \\
   & TestSGD$_{\theta=0.005}$ & 0.0143 & 0.0855 & 0.0139  & 0.0838 & 0.0077 
   
  & 0.0353 & 0.0632 & 0.0208  & 0.0065 & 0.0428  
  \\
  
  & TestSGD$_{\theta=0.002}$ & 0.1883 & 0.2240 & 0.1036 & 0.0904 & 0.0265 
  
  & 0.0428 & 0.0667 & 0.0252  & 0.0059 & \textbf{0.0652} 
  \\
    
   & \tabsysname & \textbf{0.2101} & \textbf{0.6419} & \textbf{0.2317} & \textbf{0.3892} & \textbf{0.0567} 
   
   & \textbf{0.0676} & \textbf{0.1134} & \textbf{0.0349}  & 0.0078 & 0.0615 
   \\
  \midrule
  \multirow{4}{*}{\textbf{DCN}} & Themis & 0.2110 & 0.5138 & \textbf{0.1734} & 0.2285 & 0.0482 
  
  & \textbf{0.0651} & 0.0992 & 0.0345 & \textbf{0.0072} & \textbf{0.0655} 
  
  \\
  & TestSGD$_{\theta=0.005}$ & 0.0105 & 0.0149 & 0.0094 & 0.0826 & 0.0063 
  
  & 0.0375 & 0.0572 & 0.0199 & 0.0039 & 0.0547 
  \\
  
  & TestSGD$_{\theta=0.002}$ & 0.1780 & 0.2184 & 0.0831  & 0.0996 & 0.0282  
  
  & 0.0441 & 0.0727 & 0.0267 & 0.0053 & \textbf{0.0655} 
  \\
    
  & \tabsysname & \textbf{0.2378} & \textbf{0.6113} & 0.1648 & \textbf{0.2742} & \textbf{0.0541} 
  
  & 0.0629 & \textbf{0.1018} & \textbf{0.0381} & \textbf{0.0072} & 0.0602 
  \\
  \midrule
  \multirow{4}{*}{\textbf{FGCNN}} & Themis & \textbf{0.2256} & 0.5833 & 0.2200 & 0.2682 & \textbf{0.0495} 
  
  & 0.0486 & 0.0812  & 0.0289 & \textbf{0.0088} & 0.0645 
  \\
  
  & TestSGD$_{\theta=0.005}$ & 0.0199 & 0.1034 & 0.0320  & 0.0310 & 0.0051 
  
  & 0.0380 & 0.0564 & 0.0245 & 0.0041 & 0.0476 
  \\
  
  & TestSGD$_{\theta=0.002}$ & 0.1387 & 0.2662 & 0.0881 & 0.1512 & 0.0264 
  
  & \textbf{0.0621} & 0.0812 & 0.0289 & 0.0070 & \textbf{0.0656} 
  \\

  & \tabsysname & 0.2126 & \textbf{0.7445} & \textbf{0.2615} & \textbf{0.4454} & 0.0483 
  
  & 0.0613 & \textbf{0.1012} & \textbf{0.0431} & 0.0080 & 0.0569 
  \\
  \bottomrule
\end{tabular} }
\end{table*}

\end{document}